\documentclass[letterpaper]{article} 
\usepackage{aaai2027}  
\nocopyright
\usepackage[hyphens]{url}  
\usepackage{graphicx} 
\usepackage{amsmath}
\usepackage{amssymb}
\usepackage{natbib}  
\usepackage{caption} 
\usepackage{algorithm}
\usepackage{algorithmic}

\usepackage{newfloat}
\usepackage{listings}
\usepackage{tabularx}
\DeclareCaptionStyle{ruled}{labelfont=normalfont,labelsep=colon,strut=off} 
\floatstyle{ruled}
\newfloat{listing}{tb}{lst}{}
\floatname{listing}{Listing}

\usepackage{booktabs}

\title{PRIMA: Pre-Training with Risk-Integrated Image--Metadata Alignment for Medical Diagnosis with LLM-Based Feature Aggregation}

\author {
    Yiqing Wang\textsuperscript{\rm 1},
    Chunming He\textsuperscript{\rm 1},
    Ziyun Yang\textsuperscript{\rm 1},
    Maria Woodward\textsuperscript{\rm 2},
    Ming-Chen Lu\textsuperscript{\rm 2},
    Mercy Pawar\textsuperscript{\rm 2},
    Leslie Niziol\textsuperscript{\rm 2},
    Sina Farsiu\textsuperscript{\rm 1}\corresponding
}
\affiliations {
    \textsuperscript{\rm 1}Department of Biomedical Engineering, Duke University\\
    101 Science Drive, Durham, NC 27705 USA\\
    \textsuperscript{\rm 2}Department of Ophthalmology and Visual Sciences, University of Michigan\\
    1000 Wall Street, Ann Arbor, MI, USA\\
    yq.wang@duke.edu, chunming.he@duke.edu, ziyun.yang@duke.edu, mariawoo@med.umich.edu, mingchlu@med.umich.edu, mpawar@med.umich.edu, lmniziol@med.umich.edu, sina.farsiu@duke.edu
}

\begin{document}
\frenchspacing

\maketitle

\begin{abstract}
Medical diagnosis requires the effective synthesis of visual manifestations and clinical metadata. However, existing methods often treat metadata as isolated tags, failing to exploit the rich semantic knowledge embedded in clinical descriptions. We propose PRIMA (Pre-training with Risk-integrated Image-Metadata Alignment), a framework that integrates domain-specific knowledge into multi-modal representation learning. We first curate a corpus of risk--disease correlations via Retrieval-Augmented Generation (RAG) to refine Clinical ModernBERT, embedding diagnostic priors into the text encoder. To bridge the modality gap, we introduce a dual-encoder pre-training strategy utilizing DINOv3 and our refined Clinical ModernBERT, optimized by a suite of four complementary loss functions. These losses are designed to capture multi-granular semantic alignment and handle the ambiguity of clinical correlations through soft labels. Finally, we leverage Qwen3 to fuse these aligned features for precise disease classification. Extensive experiments demonstrate that PRIMA effectively harmonizes pixel-level features with abstract clinical expertise. Across PAD-UFES-20 and AQUA, PRIMA achieves average F1-scores of 72.04\% and 85.22\%, outperforming strong image-only, metadata-fusion, and medical vision-language baselines. Notably, our framework achieves strong performance without requiring massive data collection or exhaustive computational resources. Our code is available at \url{https://github.com/yqwang01/PRIMA}.
\end{abstract}


\section{Introduction}

\begin{figure}[!tbhp]
\centering 
\includegraphics[width=0.9\linewidth]{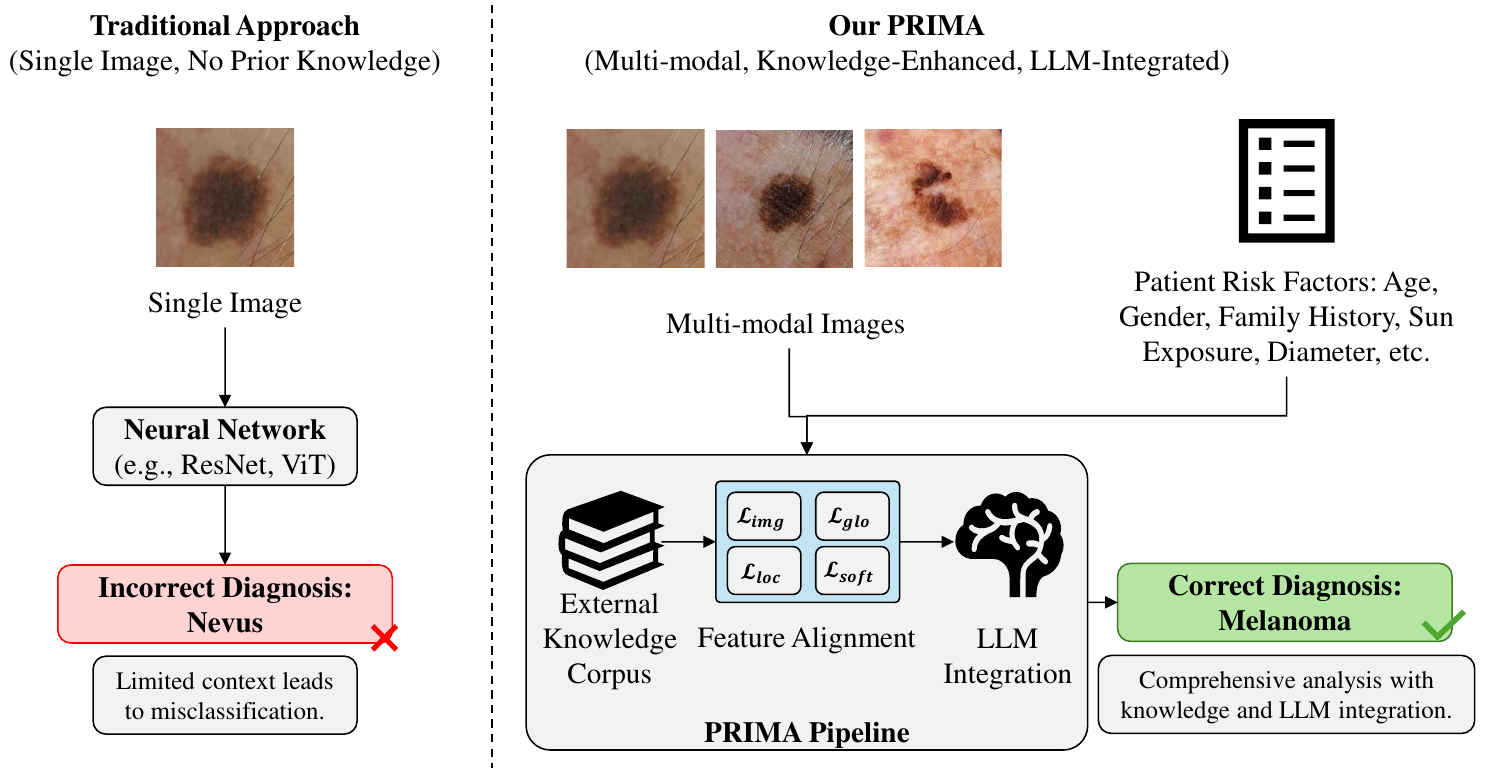}
\caption{Traditional vs. Our PRIMA Approach.}
\label{fig:intro}
\end{figure}

Medical imaging is central to clinical diagnosis, where experts synthesize information from multi-view images and complementary metadata, such as patient risk factors. While deep learning has achieved substantial success, most existing methods remain limited to single-image analysis and overlook the heterogeneous nature of real-world clinical data. This creates a gap between current algorithms and clinical diagnostic protocols, which often integrate diverse imaging evidence with structured risk profiles. Moreover, data scarcity remains a persistent barrier. Despite recent efforts to curate large-scale medical datasets~\cite{wang2025sam,li2025boosting,silva2025foundation}, relying on massive data is often infeasible for specialized tasks or rare diseases where patient cohorts are inherently limited.

Recent metadata-aware and vision-language methods provide promising directions, but they are not fully suited to the report-scarce, metadata-rich setting considered in this work. Metadata-fusion approaches~\cite{lu2023collaborative,pacheco2021attention} typically combine image and tabular features architecturally, but often lack explicit semantic modeling of risk--disease relationships. Meanwhile, CLIP-style and LLM-based medical models~\cite{radford2021learning,wang2022medclip,wang2022multi,du2024ret} usually depend on large-scale pretraining, paired reports, EHR text, or carefully curated textual annotations. As a result, the semantic potential of routinely available structured clinical metadata remains underexplored.

In this paper, we propose PRIMA (Pre-training with Risk-integrated Image-Metadata Alignment), a framework designed to integrate domain-specific clinical knowledge, patient metadata, and visual representations. PRIMA first constructs a task-specific knowledge bank of risk--disease relationships by applying Retrieval-Augmented Generation (RAG)~\cite{amugongo2025retrieval} with GPT~\cite{singh2025openai,openai2026systemcard} and Gemini~\cite{comanici2025gemini,googledeepmind2026gemini3pro} on public clinical literature. The resulting corpus is used to fine-tune Clinical ModernBERT~\cite{lee2025clinical}, injecting diagnostic priors without requiring massive paired image--text datasets. PRIMA then aligns visual features from DINOv3~\cite{simeoni2025dinov3} with the refined text encoder through multi-granular, metadata-aware objectives. Finally, Qwen3~\cite{qwen3} aggregates the aligned multi-modal features for disease prediction. Evaluations on PAD-UFES-20~\cite{Pacheco2020PADUFES20AS} and AQUA demonstrate that PRIMA consistently outperforms strong visual, metadata-fusion, and medical vision-language baselines.

\begin{figure*}[!tbhp]
\centering 
\includegraphics[width=0.8\textwidth]{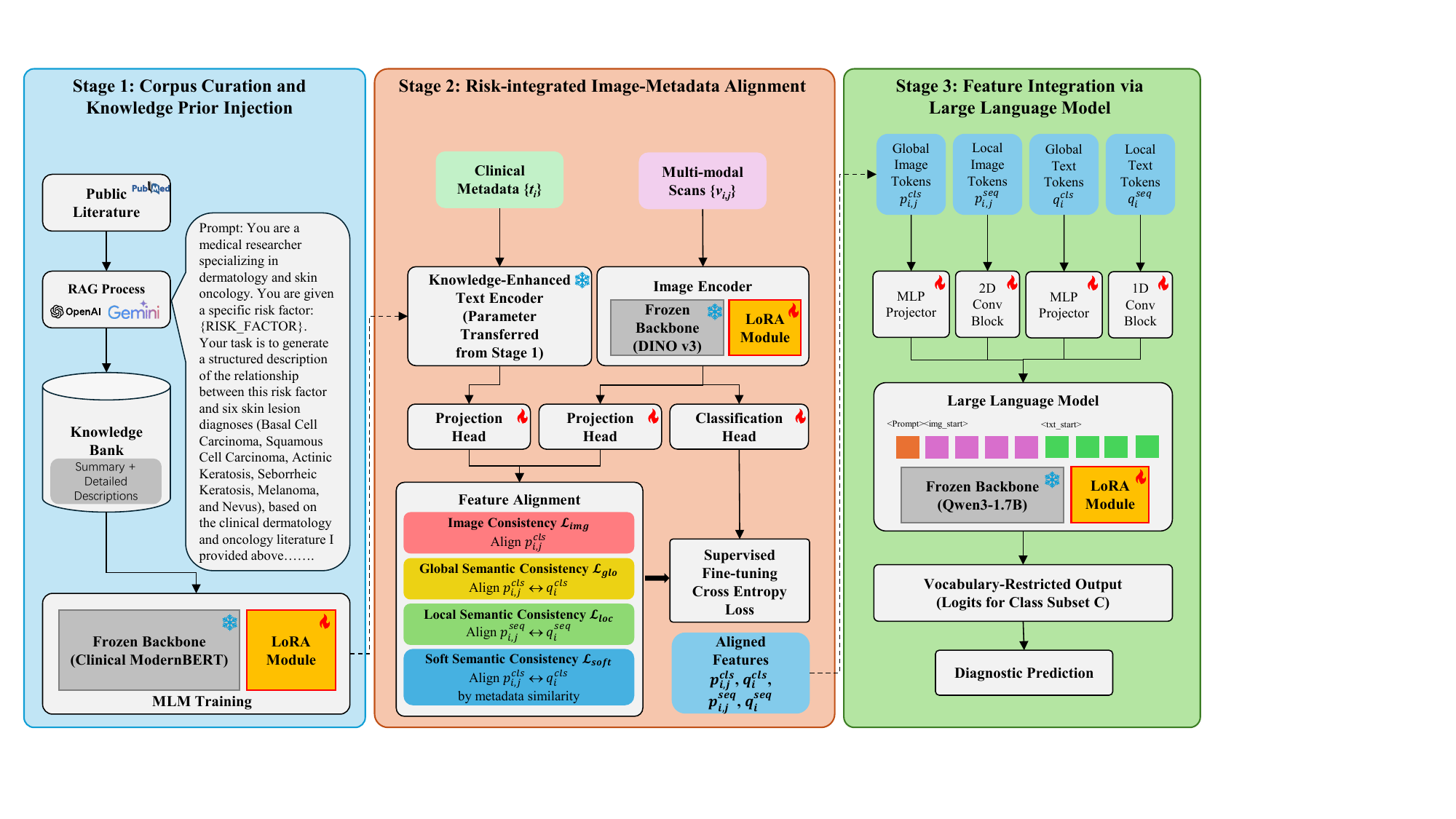}
\caption{Overview of our PRIMA.}
\label{fig:framework}
\end{figure*}

Our contributions are summarized as follows:
\begin{itemize}
    \item \textbf{Knowledge-enhanced encoding:} We elevate structured metadata into semantic knowledge by injecting RAG-derived risk--disease priors into a clinical text encoder, without requiring massive paired datasets.
    \item \textbf{Multi-granular alignment:} We introduce complementary objectives that align images, textual priors, and metadata-derived semantics at global and local levels.
    \item \textbf{LLM-driven diagnosis:} We develop a unified pipeline that uses Qwen3 to synthesize aligned multi-modal features, achieving strong performance on both public dermatology and private keratitis datasets.
\end{itemize}

\section{Related Work}

Clinical metadata is widely used to complement medical images. Prior studies, such as Pacheco and Krohling~\cite{pacheco2021attention} and Lu et al.~\cite{lu2023collaborative}, combine visual and structured features through attention mechanisms or task-specific fusion architectures. Although effective, these methods typically treat metadata as auxiliary numerical or categorical inputs. PRIMA instead uses metadata as semantic anchors, deriving patient similarities to guide representation learning and align clinical attributes with visual features through multi-granular objectives.

Medical vision-language models, including MedCLIP~\cite{wang2022medclip}, MGCA~\cite{wang2022multi}, MedKLIP~\cite{wu2023medklip}, RET-CLIP~\cite{du2024ret}, and MedBLIP~\cite{chen2024medblip}, learn representations from paired reports, EHR text, or large-scale image--text corpora. Related approaches use textual descriptors or symbolic concepts, such as FLAIR~\cite{silva2025foundation} and KnoBo~\cite{yang2024textbook}. However, these methods often require rich textual supervision, large paired datasets, or manually defined concepts. PRIMA instead targets a report-scarce but metadata-rich setting by injecting RAG-derived risk--disease priors into a clinical text encoder and using metadata as semantic anchors for multi-granular visual--textual alignment and similarity-based soft supervision.

\section{Method}

\subsection{Overview}

Fig.~\ref{fig:framework} illustrates the overall architecture of PRIMA, which comprises three progressive training stages. First, we curate a domain-specific knowledge bank via Retrieval-Augmented Generation using GPT~\cite{singh2025openai,openai2026systemcard} and Gemini~\cite{comanici2025gemini,googledeepmind2026gemini3pro} based on public literature. This corpus serves as the foundation for fine-tuning Clinical ModernBERT~\cite{lee2025clinical} to inject medical priors. Next, we introduce four complementary objectives to align visual and textual features across hierarchical semantic levels and metadata-derived relationships. Finally, a large language model is employed to synthesize the aligned multi-modal features for diagnostic prediction. In summary, Stage 1 injects domain priors into the text encoder, Stage 2 aligns image features with visually grounded metadata semantics, and Stage 3 uses all available patient attributes and aligned multi-modal features for vocabulary-restricted diagnostic classification.

\subsection{Corpus Curation and Knowledge Prior Injection}
\label{sec:knowledge}

Although LLMs provide strong general-purpose language representations, they may still struggle in medical scenarios that require fine-grained attribute reasoning, rare-disease knowledge, and domain-specific diagnostic priors. Clinical ModernBERT~\cite{lee2025clinical} partially mitigates this limitation through clinical-domain pretraining, but downstream medical datasets are often too small to support sufficient task-specific representation learning. To address this issue, we construct task-specific knowledge corpora from PubMed~\cite{lindberg2000internet} and inject the resulting diagnostic priors into the text encoder.

To avoid dataset-specific information leakage, corpus construction uses only the names of the predefined metadata fields as retrieval queries. It does not use any patient-level metadata values, images, diagnostic labels, data partitions, outcome frequencies, or dataset-level statistics from PAD-UFES-20 or AQUA. Publications describing either evaluation dataset are also excluded from retrieval. For each task, we assembled a grounding collection of approximately 25 PubMed-indexed publications covering the 21 metadata-derived factors in PAD-UFES-20 and the 37 factors in AQUA. The retrieved articles include systematic reviews, clinical studies, and case reports that describe risk--disease associations and relevant clinical attributes. 

Clinical evidence is often distributed across multiple papers, expressed implicitly, or partially inconsistent across studies. Therefore, we employ GPT~\cite{singh2025openai,openai2026systemcard} and Gemini~\cite{comanici2025gemini,googledeepmind2026gemini3pro} through Retrieval-Augmented Generation (RAG)~\cite{amugongo2025retrieval} to synthesize the retrieved literature into structured knowledge descriptions. The LLMs are used only for offline RAG-based summarization, with generated statements grounded in the retrieved sources. For each risk factor, the output contains a global summary and disease-specific paragraphs that describe clinically relevant associations, risk patterns, and diagnostic implications.

Following the Clinical ModernBERT protocol, we inject the curated knowledge into the text encoder using Masked Language Modeling (MLM)~\cite{devlin2019bert}. To preserve the pretrained representation while maintaining computational efficiency, we adopt LoRA~\cite{hu2022lora} and lightweight projection modules. Stage-specific trainable parameter ratios are reported in the supplementary material. Importantly, the curated text is used only for prior injection into the encoder, not for direct prediction or test-time reasoning. Therefore, residual noise or uncertainty in the synthesized knowledge is unlikely to directly determine the final diagnosis, but instead provides weak domain priors that guide the subsequent multi-granular alignment process.

\subsection{Risk-integrated Image-Metadata Alignment}
\label{sec:alignment}

\subsubsection{Image and Text Encoder} We employ DINOv3~\cite{simeoni2025dinov3} and our knowledge-enhanced Clinical ModernBERT~\cite{lee2025clinical} as the vision and text backbones. For the $i$-th instance, the inputs consist of multi-modal images $\{v_{i,j}\}_j$ and structured clinical metadata $\{t_{i,k}\}_k$. The encoders extract textual features $q_i\in\mathbb{R}^{(L+1)\times d}$ and visual features $p_{i,j}\in\mathbb{R}^{(K+1)\times d}$ from each image, where $L$ is the number of textual sequence tokens, $K$ is the number of visual patch tokens, and $d$ is the latent feature dimension. The additional token in each sequence denotes the corresponding global class token. Both modalities are mapped into a shared representation space through projection heads. At this stage, only the projection heads and the LoRA~\cite{hu2022lora} adapters in DINOv3 are optimized (Fig.~\ref{fig:framework}).

For Stage~2 image--metadata alignment, we retain only attributes whose semantics are directly related to visually observable findings, thereby reducing noise from metadata that cannot be grounded in image features. The attributes are selected through a predefined semantic grouping of the metadata fields rather than learned by the model; for example, PAD-UFES-20 retains lesion-related attributes such as itch, growth, pain, and change, while excluding demographic attributes such as age and gender. This provides a more appropriate supervision signal for visual--semantic alignment. The filtering is applied only in Stage~2, whereas all available patient attributes are used in Stage~3 for diagnostic feature aggregation and classification. The complete attribute lists for both datasets are provided in the supplementary material.

\subsubsection{Alignment Strategy} 
\begin{figure*}[!tbhp]
\centering 
\includegraphics[width=0.75\textwidth]{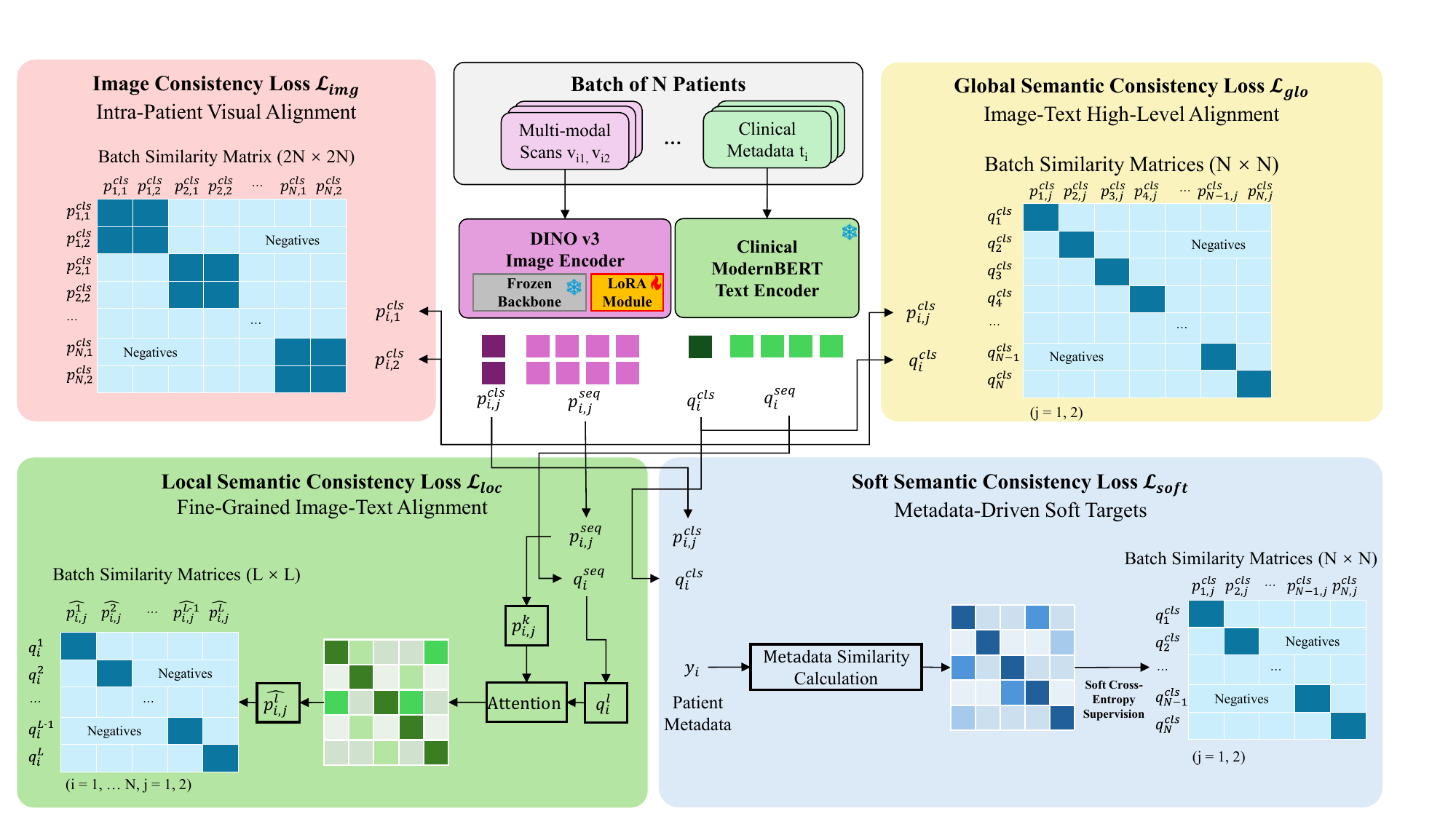}
\caption{Training Objectives of Our PRIMA.}
\label{fig:loss}
\end{figure*}

We decompose $p_{i,j}$ and $q_i$ into global class tokens ($p_{i,j}^{cls}, q_i^{cls}$) and local sequence tokens ($p_{i,j}^{seq}, q_i^{seq}$) to facilitate multi-granular alignment through four complementary objectives (Fig.~\ref{fig:loss}). Image Consistency Loss ($\mathcal{L}_{img}$) enforces same-instance visual consistency by aligning global visual features across images. Global Semantic Loss ($\mathcal{L}_{glo}$) synchronizes visual and textual class tokens for high-level semantic alignment, while Local Semantic Loss ($\mathcal{L}_{loc}$) captures fine-grained correlations between image patches and textual tokens. To handle clinical ambiguity, Soft Semantic Loss ($\mathcal{L}_{soft}$) provides soft supervision via metadata-based similarity matrices. The final objective is a weighted sum of these losses. Following alignment, the image encoder undergoes supervised fine-tuning with ground-truth labels to sharpen its discriminative power.

\textbf{Image Consistency Loss $\mathcal{L}_{img}$: } To promote visual consistency across views of the same training instance, we construct each positive pair using two distinct images from the same instance or, when only one image is available, two independently augmented copies of that image. Here, a training instance corresponds to a lesion in PAD-UFES-20 and a subject in AQUA. The vision encoder extracts a global class token from each view, and their similarity is measured using the temperature-scaled cosine similarity in Eq.~\ref{eq:sim}, with temperature hyperparameter $\tau$.
{\footnotesize
    \begin{equation}
    \label{eq:sim}
    \operatorname{s}(x_1,x_2)
    =
    \frac{x_1^\top x_2}
    {\tau\lVert x_1\rVert_2\lVert x_2\rVert_2}.
    \end{equation}}
For a batch of $N$ instances, let $p_{i,1}^{cls}$ and $p_{i,2}^{cls}$ be global visual tokens from two views (distinct images or augmentations) of patient $i$. Defined in Eq.~\ref{eq:img}, $\mathcal{L}_{img}$ promotes view-invariant representation learning by aligning latent features belonging to the same subject.
{\footnotesize
    \begin{align}
    \mathcal{L}_{\mathrm{img}}
    &=
    -\frac{1}{2N}
    \sum_{i=1}^{N}
    \sum_{\substack{a,b\in\{1,2\}\\a\neq b}}
    \Bigg[
    s\!\left(p_{i,a}^{\mathrm{cls}},p_{i,b}^{\mathrm{cls}}\right)
    \notag\\
    &\quad-
    \log\Bigg(
    \sum_{k=1}^{N}
    e^{s(p_{i,a}^{\mathrm{cls}},p_{k,b}^{\mathrm{cls}})}
    +
    \sum_{\substack{k=1\\k\neq i}}^{N}
    e^{s(p_{i,a}^{\mathrm{cls}},p_{k,a}^{\mathrm{cls}})}
    \Bigg)
    \Bigg].
    \label{eq:img}
    \end{align}
    }

\textbf{Global Semantic Loss $\mathcal{L}_{glo}$: }To establish a shared feature space, the Global Semantic Loss ($\mathcal{L}_{glo}$) employs a symmetric cross-entropy objective (Eq. \ref{eq:glo}) over $N$ matched global pairs $\{(p_{i,j}^{cls}, q_i^{cls})\}_{i=1,j=1}^{N,2}$. This synchronizes high-level semantic context, ensuring global visual embeddings capture the abstract clinical concepts encoded in the metadata.
{\footnotesize
\begin{align}
\mathcal{L}_{\mathrm{glo}}
&=
-\frac{1}{4N}\sum_{i=1}^{N}\sum_{j=1}^{2}
\Bigg[
\log
\frac{
e^{s(p_{i,j}^{\mathrm{cls}},q_i^{\mathrm{cls}})}
}{
\sum_{k=1}^{N}
e^{s(p_{i,j}^{\mathrm{cls}},q_k^{\mathrm{cls}})}
}
\notag\\
&\qquad\qquad
+
\log
\frac{
e^{s(q_i^{\mathrm{cls}},p_{i,j}^{\mathrm{cls}})}
}{
\sum_{k=1}^{N}
e^{s(q_i^{\mathrm{cls}},p_{k,j}^{\mathrm{cls}})}
}
\Bigg].
\label{eq:glo}
\end{align}
}

\textbf{Local Semantic Loss $\mathcal{L}_{loc}$: }
To capture fine-grained image--text correlations, we construct a token-conditioned visual representation $\hat{p}_{i,j}^l$ for each text token $q_i^l$ in $q_i^{\mathrm{seq}}=\{q_i^l\}_{l=1}^{L}$. Specifically, $q_i^l$ attends to the $K$ patch tokens $p_{i,j}^{\mathrm{seq}}=\{p_{i,j}^k\}_{k=1}^{K}$ from visual view $j$, and the patches are aggregated according to their relevance to the token:
{\footnotesize
\begin{align}
\alpha_{i,j}^{l,k}
&=
\frac{
\exp\!\left((q_i^l)^\top p_{i,j}^k/\sqrt{d}\right)
}{
\sum_{r=1}^{K}
\exp\!\left((q_i^l)^\top p_{i,j}^r/\sqrt{d}\right)
},
\hat{p}_{i,j}^l
=
\sum_{k=1}^{K}
\alpha_{i,j}^{l,k}p_{i,j}^k.
\label{eq:attn}
\end{align}
}

We then form an $L\times L$ similarity matrix between the text tokens and their token-conditioned visual representations. The loss performs symmetric matching: the text-to-visual direction retrieves the corresponding visual representation for each text-token anchor, while the visual-to-text direction retrieves the corresponding text token for each visual anchor. Averaged over both visual views, the loss is
{\footnotesize
\begin{align}
\mathcal{L}_{\mathrm{loc}}
&=
-\frac{1}{4LN}
\sum_{i=1}^{N}
\sum_{j=1}^{2}
\sum_{l=1}^{L}
\Bigg[
\log
\frac{
e^{s(q_i^l,\hat{p}_{i,j}^l)}
}{
\sum_{m=1}^{L}
e^{s(q_i^l,\hat{p}_{i,j}^m)}
}
\notag\\
&\qquad\qquad+
\log
\frac{
e^{s(\hat{p}_{i,j}^l,q_i^l)}
}{
\sum_{m=1}^{L}
e^{s(\hat{p}_{i,j}^l,q_i^m)}
}
\Bigg].
\label{eq:loc}
\end{align}
}

\textbf{Soft Semantic Loss $\mathcal{L}_{soft}$: } Standard contrastive learning assumes one-to-one image--text correspondence and does not account for shared clinical characteristics across samples. We therefore construct a multi-hot semantic vector $y_i$ from the diagnostic label and binary metadata attributes. The diagnostic-label dimensions are weighted by 3, while the metadata dimensions retain unit weight, and $y_i$ is $\ell_2$-normalized before similarity computation. The complete attributes used are provided in the supplementary material. Let $\omega_{ij}=\operatorname{Softmax}_{j}\!\left(\langle y_i,y_j\rangle/\tau_{\mathrm{label}}\right)$ denote the image-to-text soft target and $\widetilde{\omega}_{ji}=\operatorname{Softmax}_{i}\!\left(\langle y_i,y_j\rangle/\tau_{\mathrm{label}}\right)$ denote the text-to-image target, where $\tau_{\mathrm{label}}$ controls target sharpness (default 0.05). Because the same semantic vectors are used on both sides, $\langle y_i,y_j\rangle$ is symmetric and $\widetilde{\omega}_{ji}=\omega_{ji}$. We then minimize the bidirectional soft cross-entropy between these targets and the predicted image--text similarities:

{\footnotesize
\begin{align}
\mathcal{L}_{\mathrm{soft}}
&=
-\frac{1}{4N}
\sum_{i=1}^{N}\sum_{j=1}^{N}\sum_{k=1}^{2}
\Bigg[
\omega_{ij}
\log
\frac{
e^{s(p_{i,k}^{\mathrm{cls}},q_j^{\mathrm{cls}})}
}{
\sum_{l=1}^{N}
e^{s(p_{i,k}^{\mathrm{cls}},q_l^{\mathrm{cls}})}
}
\notag\\
&\qquad\qquad
+
\widetilde{\omega}_{ji}
\log
\frac{
e^{s(q_j^{\mathrm{cls}},p_{i,k}^{\mathrm{cls}})}
}{
\sum_{l=1}^{N}
e^{s(q_j^{\mathrm{cls}},p_{l,k}^{\mathrm{cls}})}
}
\Bigg].
\label{eq:soft}
\end{align}
}

The final alignment loss is as shown in Eq. \ref{eq:final}. The weight coefficients $\beta_{1}$, $\beta_{2}$, $\beta_{3}$ and $\beta_{4}$ are set as $\beta_{1} = 0.2, \beta_{2}=0.3, \beta_{3}=0.2, \beta_{4}=0.3$.
\begin{equation}
\label{eq:final}
\mathcal{L}_{\mathrm{align}}
=
\beta_1\mathcal{L}_{\mathrm{img}}
+\beta_2\mathcal{L}_{\mathrm{glo}}
+\beta_3\mathcal{L}_{\mathrm{loc}}
+\beta_4\mathcal{L}_{\mathrm{soft}}.
\end{equation}
Following feature alignment, the image backbone undergoes supervised refinement using ground-truth labels and Cross-Entropy loss \cite{mao2023cross}. This stage sharpens visual representations by bridging the gap between general cross-modal alignment and task-specific diagnostic requirements.

\subsection{Feature Integration via Large Language Model}
\label{sec:integration}

\begin{table*}[t]
\centering
\small
\setlength{\tabcolsep}{2.7pt}
\begin{tabular}{lccccccccc}
\toprule
\textbf{Methods} & BCC & SCC & MEL & ACK & SEK & NEV & Avg F1 & Acc & BAcc\\
\midrule
DINOv3 LoRA \cite{simeoni2025dinov3}
& 71.11 & 32.95 & 56.06 & 79.95 & 73.64 & 76.26
& 65.00 (4.33) & 70.49 (2.81) & 67.07 (1.92) \\

+ Metadata via MLP
& 76.24 & 33.56 & 66.60 & 83.43 & 72.86 & 75.91
& 68.10 (3.20) & 74.14 (2.38) & 69.21 (2.61) \\

\midrule
MedKLIP \textcolor{gray}{[ICCV-23]} \cite{wu2023medklip}
& 58.44 & 18.26 & 31.56 & 65.57 & 53.96 & 56.73
& 47.42 (1.64) & 54.89 (3.52) & 51.43 (1.44) \\

KnoBo \textcolor{gray}{[NIPS-24]} \cite{yang2024textbook}
& 57.50 & 22.51 & 26.21 & 71.30 & 42.56 & 59.52
& 46.60 (4.96) & 56.69 (8.54) & 48.07 (3.21) \\

MedBLIP \textcolor{gray}{[ACCV-24]} \cite{chen2024medblip}
& 72.10 & 33.36 & 72.08 & 82.08 & \textbf{77.20}
& \textbf{80.47}
& 69.55 (1.25) & 73.14 (1.84) & 71.90 (2.06) \\

MLRG \textcolor{gray}{[CVPR-25]} \cite{liu2025enhanced}
& 65.98 & 23.59 & 67.93 & 74.19 & 59.67 & 70.27
& 60.27 (5.53) & 64.41 (6.83) & 60.09 (4.30) \\

\midrule
\textbf{PRIMA (Ours)}
& \textbf{79.80}
& \textbf{36.84}
& \textbf{75.17}
& \textbf{84.82}
& 76.30
& 79.28
& \textbf{72.04} (3.21)
& \textbf{77.26} (1.91)
& \textbf{71.95} (3.49) \\
\bottomrule
\end{tabular}
\caption{Quantitative results on PAD-UFES-20 (\%). Values are reported as
mean (standard deviation) over five folds.}
\label{tab:pad}
\end{table*}

To leverage the feature-integration capacity of large language models, we bridge the gap between encoders and the LLM backbone via a multi-modal projection strategy. Global tokens are projected using MLP layers, while local sequence tokens are aligned through 1D/2D convolution blocks that perform spatial and sequence downsampling to reduce overhead. These features are concatenated into an input sequence demarcated by learnable special tokens (e.g., \texttt{<|img\_start|>}), which are initialized with semantic embeddings to accelerate convergence. To maintain efficiency and prevent overfitting, we utilize LoRA \cite{hu2022lora}, updating only the projectors and 2.04\% of total parameters. We further employ a vocabulary-restricted strategy to avoid free-form generation and reduce hallucination risk: instead of free-form generation, we extract logits $z_k$ exclusively from a token subset $\mathcal{C}$ corresponding to pre-defined clinical classes. Probabilities are computed via Softmax over this constrained set: $P(y=k|x) = \frac{\exp(z_k)}{\sum_{j \in \mathcal{C}} \exp(z_j)}$. The model is optimized by Cross Entropy loss \cite{mao2023cross}. Unlike Stage 2, which filters metadata to retain visually grounded attributes for alignment, Stage 3 uses all available patient attributes together with the aligned visual and textual features for final diagnostic aggregation.

\section{Experiments and Results}

\subsection{Dataset Details}

We evaluated PRIMA on PAD-UFES-20~\cite{Pacheco2020PADUFES20AS} and AQUA. PAD-UFES-20 contains 2,298 images from 1,373 patients across six skin-lesion categories. AQUA contains 19,567 slit-lamp images from 1,827 subjects for bacterial and fungal keratitis diagnosis, with up to three illumination modalities per case collected from India and US. Detailed cohort characteristics, label-confirmation procedures, metadata attributes, and class distributions are provided in the supplementary material. We used patient-level five-fold cross-validation with same-fold checkpoint selection for both datasets; no patient appeared in multiple folds, and no external test set was used. All methods used identical folds and preprocessing pipelines. We report the mean and standard deviation of F1-score, accuracy, and balanced accuracy across folds. We used fixed random seeds to improve reproducibility.

\subsection{Implementation Details}

PRIMA was trained on two NVIDIA RTX 4090 GPUs using AdamW with warm-up. The learning rates for Stages 1--3 were $5\times10^{-5}$, $3\times10^{-5}$, and $1\times10^{-5}$, respectively. LoRA was used for parameter-efficient fine-tuning. Images were resized to $224\times224$, the projection dimension was 768, and the maximum text length was 512 tokens. Complete optimization, augmentation, architectural, and reproducibility details are provided in the supplementary material. Our code is available at \url{https://github.com/yqwang01/PRIMA}.

\subsection{Main Results}

\begin{table*}[t]
\centering
\small
{
\setlength{\tabcolsep}{5pt}
\begin{tabular}{lccccc|ccc|c}
\toprule
&
\multicolumn{5}{c|}{\textbf{F1-Score}} &
\multicolumn{3}{c|}{\textbf{Accuracy}} &
\textbf{BAcc.} \\
\textbf{Method}
& Fungal
& Bacterial
& India
& US
& Overall
& India
& US
& Overall
& Overall \\
\midrule

DINOv3 LoRA (ScS) \cite{simeoni2025dinov3}
& 82.54
& 71.42
& 67.61
& 67.40
& 76.98 (1.78)
& 77.69
& 80.15
& 78.37 (1.60)
& 78.25 (1.42) \\

Multi-modal via MLP (w/o metadata)
& 83.99
& 76.54
& 69.49
& 70.91
& 80.26 (3.84)
& 79.73
& 83.67
& 81.00 (3.83)
& 81.00 (3.09) \\

+ Metadata via MLP
& 86.36
& 79.10
& 72.11
& 71.75
& 82.73 (2.63)
& 82.96
& 84.81
& 83.52 (2.66)
& 83.15 (1.93) \\

\midrule

DeepIK \textcolor{gray}{[NPJ Dig. Med. - 24]}\cite{li2024deep}
& 80.84
& 76.54
& 68.08
& 71.75
& 78.69 (3.00)
& 75.46
& 86.51
& 78.93 (2.94)
& 81.11 (2.38) \\

KnoBo \textcolor{gray}{[NIPS-24]}\cite{yang2024textbook}
& 83.54
& 72.42
& 58.80
& 54.31
& 77.98 (6.09)
& 79.70
& 78.61
& 79.42 (5.72)
& 77.76 (5.75) \\

MedBLIP \textcolor{gray}{[ACCV-24]}\cite{chen2024medblip}
& 80.96
& 75.55
& 70.40
& 64.96
& 78.26 (5.13)
& 77.73
& 80.19
& 78.82 (5.37)
& 79.86 (4.34) \\

MLRG \textcolor{gray}{[CVPR-25]}\cite{liu2025enhanced}
& 83.35
& 74.28
& 65.45
& 68.20
& 78.81 (2.73)
& 78.78
& 82.23
& 79.91 (2.44)
& 79.15 (2.81) \\

\midrule

\textbf{PRIMA (Ours)}
& \textbf{88.66}
& \textbf{81.78}
& \textbf{73.25}
& \textbf{74.39}
& \textbf{85.22} (2.17)
& \textbf{85.23}
& \textbf{87.77}
& \textbf{86.04} (1.82)
& \textbf{85.40} (2.27) \\

\bottomrule
\end{tabular}
}
\caption{Quantitative results on the AQUA dataset (\%). Values are reported as mean (standard deviation) over five folds.}
\label{tab:aqua}
\end{table*}

We evaluated PRIMA against representative baselines on PAD-UFES-20 and AQUA. We first fine-tuned DINOv3~\cite{simeoni2025dinov3} with LoRA as a strong pure-image baseline, and further combined its visual features with vanilla Clinical ModernBERT~\cite{lee2025clinical} embeddings through an MLP to assess simple metadata fusion. We also compared PRIMA with prominent medical diagnostic baselines, including MedKLIP~\cite{wu2023medklip}, KnoBo~\cite{yang2024textbook}, MedBLIP~\cite{chen2024medblip}, MLRG~\cite{liu2025enhanced}, and DeepIK~\cite{li2024deep}.

As shown in Table~\ref{tab:pad}, DINOv3 LoRA achieved a 65.00\% average F1-score and 70.49\% accuracy on PAD-UFES-20, while metadata fusion improved them to 68.10\% and 74.14\%, respectively. PRIMA further increased the average F1-score to 72.04\% and accuracy to 77.26\%, exceeding the image-only baseline by 7.04 and 6.77 percentage points and the input-matched metadata-fusion baseline by 3.94 and 3.12 points, respectively. PRIMA also achieved the best F1-score for BCC, SCC, MEL, and ACK. Several medical vision--language baselines performed below DINOv3, likely because they rely on large-scale medical pretraining and specialized textual annotations that are not directly available or well matched in this setting. This suggests that strong generic visual representations may transfer better than report-based medical VLMs when paired reports or task-matched textual annotations are unavailable. Although SCC remains challenging because of class imbalance, PRIMA improves its F1-score to 36.84\%, outperforming DINOv3, metadata fusion, and MedBLIP by 3.89, 3.28, and 3.48 percentage points, respectively. PRIMA was higher than all six comparison methods in accuracy on all five reported folds (One-sided Wilcoxon signed-rank tests, p=0.03125).

Table~\ref{tab:aqua} reports the AQUA results. The single-modality DINOv3 LoRA baseline using sclerotic-scatter imaging achieved a 76.98\% average F1-score and 78.37\% accuracy. Fusing all three imaging modalities with an MLP increased these metrics to 80.26\% and 81.00\%, while further incorporating metadata achieved 82.73\% and 83.52\%. PRIMA achieved the best overall average F1-score, accuracy, and balanced accuracy of 85.22\%, 86.04\%, and 85.40\%, exceeding the input-matched three-modality image--metadata MLP by 2.49, 2.52, and 2.25 percentage points, respectively. PRIMA also improved the site-specific macro F1-score from 72.11\% to 73.25\% in India and from 71.75\% to 74.39\% in the US. Nevertheless, performance across different sites is substantially lower than the pooled results, suggesting the presence of location bias in the learned representations; we further discuss this in the supplementary material. PRIMA was higher on all five reported folds than DINOv3, metadata fusion, DeepIK, KnoBo, and MLRG for the three overall metrics (One-sided Wilcoxon signed-rank tests, p=0.03125).

Overall, PRIMA consistently outperforms pure-image learning, simple metadata fusion, specialized diagnostic networks, and recent LLM-based medical models across two distinct datasets. These results suggest that directly transferring existing medical vision-language models can be suboptimal when large-scale domain-specific pretraining data or carefully curated textual annotations are unavailable. PRIMA addresses this limitation by using LLMs for offline knowledge extraction and aligning visual, textual, and metadata-derived semantics through task-specific training.


\subsection{Ablation Study}

We conduct ablation studies on PAD-UFES-20 to evaluate PRIMA from multiple perspectives. Additional analyses are provided in the supplementary material.

\subsubsection{Effect of Multi-granular Semantic Alignment}

Table~\ref{tab:ablation_loss} reports the progressive effects of the training objectives without knowledge pretraining. The baseline denotes the PRIMA pipeline trained only with class-level supervision, without knowledge pretraining or the proposed alignment losses, and therefore differs from the standalone DINOv3 LoRA baseline in Table~\ref{tab:pad}. It achieves 66.73\% average F1-score, 72.82\% accuracy, and 65.88\% balanced accuracy, indicating that LLM-based aggregation alone cannot fully exploit limited medical supervision. Adding $\mathcal{L}_{\mathrm{img}}$ increases accuracy to 74.88\%, suggesting improved view-level stability, although the slight F1-score decrease shows that visual consistency alone is insufficient for fine-grained discrimination. By contrast, $\mathcal{L}_{\mathrm{glo}}$ raises the average F1-score to 67.82\% and balanced accuracy to 68.66\%, demonstrating the benefit of global image--text alignment. Combining $\mathcal{L}_{\mathrm{img}}$ and $\mathcal{L}_{\mathrm{glo}}$ further increases accuracy to 74.93\%, but yields only marginal gains in average F1-score and balanced accuracy over the baseline, both remaining below $\mathcal{L}_{\mathrm{glo}}$ alone. This indicates that view consistency and global alignment are insufficient for balanced class discrimination without explicit local correspondence. Adding $\mathcal{L}_{\mathrm{loc}}$ increases the average F1-score to 69.02\% and balanced accuracy to 69.56\%, confirming the value of fine-grained image-region--token alignment. Finally, $\mathcal{L}_{\mathrm{soft}}$ further improves these metrics to 70.20\% and 71.10\%, respectively, showing that metadata-derived soft relationships provide supervision beyond hard labels by avoiding the assumption that all samples from different classes are equally negative.

\begin{table}[t]
\centering
\small
{
\setlength{\tabcolsep}{3.2pt}
\begin{tabular}{ccccccc}
\toprule
$\mathcal{L}_{\mathrm{img}}$
& $\mathcal{L}_{\mathrm{glo}}$
& $\mathcal{L}_{\mathrm{loc}}$
& $\mathcal{L}_{\mathrm{soft}}$
& \textbf{Avg. F1}
& \textbf{Acc.}
& \textbf{BAcc.} \\
\midrule
-- & -- & -- & --
& 66.73 & 72.82 & 65.88 \\

$\checkmark$ & -- & -- & --
& 65.89 & 74.88 & 66.33 \\

-- & $\checkmark$ & -- & --
& 67.82 & 73.51 & 68.66 \\

$\checkmark$ & $\checkmark$ & -- & --
& 66.75 & 74.93 & 65.97 \\

$\checkmark$ & $\checkmark$ & $\checkmark$ & --
& 69.02 & \textbf{75.25} & 69.56 \\

$\checkmark$ & $\checkmark$ & $\checkmark$ & $\checkmark$
& \textbf{70.20} & 74.93 & \textbf{71.10} \\
\bottomrule
\end{tabular}
}
\caption{Ablation of the multi-granular semantic losses on
PAD-UFES-20 (\%).}
\label{tab:ablation_loss}
\end{table}

\subsubsection{Effect of Loss Design} 

We further compare the proposed loss designs with simplified alternatives in Table~\ref{tab:ablation_design}. For local semantic alignment, we replace the attention-based local semantic loss with a direct local matching variant, denoted as $\mathcal{L}_{loc\_dir}$. This variant achieves an average F1-score of 69.27\%, which is lower than the full PRIMA model. The degradation indicates that direct local matching is less effective than attention-based region--token alignment, likely because clinically relevant image regions do not always have a fixed one-to-one correspondence with textual tokens.

For soft semantic supervision, we compare $\mathcal{L}_{soft}$ with a standard supervised contrastive formulation using only class labels, denoted as $\mathcal{L}_{sup\_con}$. This variant achieves 70.57\% average F1-score and 76.25\% accuracy. Although it obtains the highest balanced accuracy, it remains lower than the full model in average F1-score and accuracy. This result suggests that class labels provide useful supervision, but metadata-aware soft relationships offer a richer training signal for overall classification performance.

\begin{table}[t]
\centering
\small
{
\setlength{\tabcolsep}{4pt}
\begin{tabular}{lccc}
\toprule
\textbf{Loss Design}
& \textbf{Avg. F1}
& \textbf{Acc.}
& \textbf{BAcc.} \\
\midrule
Direct local matching
$\mathcal{L}_{\mathrm{loc\text{-}dir}}$
& 69.27
& 76.15
& 68.69 \\

Supervised contrastive
$\mathcal{L}_{\mathrm{sup\text{-}con}}$
& 70.57
& 76.25
& \textbf{72.47} \\

\textbf{PRIMA (Ours)}
& \textbf{72.04}
& \textbf{77.26}
& 71.95 \\
\bottomrule
\end{tabular}
}
\caption{Ablation of alternative loss design choices on PAD-UFES-20
(\%).}
\label{tab:ablation_design}
\end{table}

\subsubsection{Effect of Knowledge Pretraining}

Table~\ref{tab:ablation_knowledge} isolates the contribution of domain-specific knowledge pretraining. Without knowledge pretraining, the model using all proposed semantic losses achieves 70.20\% average F1-score, 74.93\% accuracy, and 71.10\% balanced accuracy. After incorporating knowledge pretraining, PRIMA improves the average F1-score to 72.04\% and accuracy to 77.26\%, while maintaining competitive balanced accuracy. These improvements demonstrate that domain-specific knowledge provides complementary clinical priors to the multi-granular semantic alignment losses. In particular, knowledge pretraining helps the model better interpret disease-related concepts and reduces the gap between general visual-language representations and domain-specific medical classification.

\begin{table}[t]
\centering
\small
{
\setlength{\tabcolsep}{4pt}
\begin{tabular}{lccc}
\toprule
\textbf{Method}
& \textbf{Avg. F1}
& \textbf{Acc.}
& \textbf{BAcc.} \\
\midrule
w/o knowledge pretraining
& 70.20
& 74.93
& 71.10 \\

\textbf{PRIMA (Ours)}
& \textbf{72.04}
& \textbf{77.26}
& \textbf{71.95} \\
\bottomrule
\end{tabular}
}
\caption{Effect of domain-specific knowledge pretraining on
PAD-UFES-20 (\%).}
\label{tab:ablation_knowledge}
\end{table}

\subsubsection{Effect of Image Encoder Choice}

We further evaluate whether the proposed alignment strategy depends on a specific visual backbone. As shown in Table~\ref{tab:ablation_backbone}, PRIMA consistently improves the average F1-score and accuracy across different image encoders on PAD-UFES-20. With a standard ViT-Base backbone~\cite{dosovitskiy2020image}, PRIMA improves the average F1-score from 60.69\% to 66.79\% and accuracy from 66.42\% to 74.62\%. With EVA ViT~\cite{fang2023eva}, PRIMA improves the average F1-score from 67.92\% to 70.92\% and accuracy from 70.76\% to 78.32\%, achieving the highest accuracy among all backbone settings. With DINOv3~\cite{simeoni2025dinov3}, PRIMA improves the average F1-score from 65.00\% to 72.04\% and accuracy from 70.49\% to 77.26\%. These results indicate that PRIMA is not backbone-specific; its main contribution lies in aligning image representations with metadata-derived semantic priors. We use DINOv3 as the default backbone because it achieves the best average F1-score and balanced accuracy after PRIMA alignment, while noting that the proposed alignment strategy also enhances other pure-image encoders.

\begin{table}[t]
\centering
\small
{
\setlength{\tabcolsep}{4pt}
\begin{tabular}{lcccc}
\toprule
\textbf{Backbone}
& \textbf{PRIMA}
& \textbf{Avg. F1}
& \textbf{Acc.}
& \textbf{BAcc.} \\
\midrule
ViT-Base
& --
& 60.69
& 66.42
& 63.89 \\

ViT-Base
& $\checkmark$
& 66.79
& 74.62
& 67.30 \\

\midrule
EVA ViT
& --
& 67.92
& 70.76
& 71.80 \\

EVA ViT
& $\checkmark$
& 70.92
& \textbf{78.32}
& 70.89 \\

\midrule
DINOv3
& --
& 65.00
& 70.49
& 67.07 \\

DINOv3
& $\checkmark$
& \textbf{72.04}
& 77.26
& \textbf{71.95} \\
\bottomrule
\end{tabular}
}
\caption{Effect of image encoder choice and PRIMA alignment on
PAD-UFES-20 (\%).}
\label{tab:ablation_backbone}
\end{table}

\subsubsection{Effect of Stage-3 Diagnostic Aggregation}

We further examine the Stage-3 diagnostic aggregation module. As shown in Table~\ref{tab:ablation_stage3}, the metadata-only model achieves 22.16\% average F1-score, 30.88\% accuracy, and 38.35\% balanced accuracy, substantially below the image-only DINOv3 LoRA results of 65.00\%, 70.49\%, and 67.07\%, respectively. This indicates that visual information provides the primary diagnostic signal, while metadata contributes complementary cues when aligned with image features. Using the same Stage-2 aligned features, Qwen3 achieves the best average F1-score and accuracy of 72.04\% and 77.26\%, exceeding the image-only baseline by 7.04 and 6.77 percentage points. It also outperforms the MLP head by 1.41 and 3.49 points and the Transformer head by 1.33 and 2.96 points in average F1-score and accuracy, respectively. Although the MLP head yields slightly higher balanced accuracy, potentially by improving class recall at the expense of precision and overall correctness, we select Qwen3 primarily based on average F1-score, with accuracy and balanced accuracy as complementary metrics. The sensitivity analysis further shows that alternative settings can improve balanced accuracy, whereas the final configuration prioritizes average F1-score and accuracy.

\begin{table}[t]
\centering
\scriptsize
{
\setlength{\tabcolsep}{3.2pt}
\begin{tabular}{llccc}
\toprule
\textbf{Input}
& \textbf{Aggregator}
& \textbf{Avg. F1}
& \textbf{Acc.}
& \textbf{BAcc.} \\
\midrule
Metadata only
& Clinical ModernBERT
& 22.16
& 30.88
& 38.35 \\

Image only
& DINOv3 LoRA
& 65.00
& 70.49
& 67.07 \\

\midrule
Stage-2 aligned features
& MLP
& 70.63
& 73.77
& \textbf{72.31} \\

Stage-2 aligned features
& Transformer
& 70.71
& 74.30
& 71.86 \\

Stage-2 aligned features
& Qwen3
& \textbf{72.04}
& \textbf{77.26}
& 71.95 \\
\bottomrule
\end{tabular}
}
\caption{Effect of Stage-3 diagnostic aggregation on PAD-UFES-20 (\%).}
\label{tab:ablation_stage3}
\end{table}

\section{Conclusion}

We introduce PRIMA, which combines medically pretrained Clinical ModernBERT with DINOv3 and four multi-granular objectives to align diagnostic priors with visual representations. A vocabulary-restricted Qwen3 classifier then aggregates the aligned image and metadata features for diagnosis. Across PAD-UFES-20 and AQUA, PRIMA consistently outperforms image-only, conventional metadata-fusion, and medical vision--language baselines through parameter-efficient adaptation, demonstrating that structured clinical metadata can serve as both auxiliary inputs and semantic anchors for image alignment. Nevertheless, internal validation may yield optimistic estimates, and external testing remains necessary. Future work will address site- and race-related fairness, clinically review the generated knowledge bank, and extend PRIMA to additional datasets and clinical settings using more effectively curated public data and knowledge bases.

\bibliography{aaai2027}

\newpage

This supplementary document provides dataset details, implementation settings, additional experiments, and statistical analyses supporting the corresponding sections of the main paper.

\section{Datasets and Experimental Protocol}
\label{sec:sup_datasets}

\subsection{PAD-UFES-20}
\label{sec:sup_pad}

PAD-UFES-20~\cite{Pacheco2020PADUFES20AS} contains 2,298 clinical images of 1,891 skin lesions collected from 1,373 patients. The dataset covers six diagnostic categories: Basal Cell Carcinoma (BCC), Squamous Cell Carcinoma (SCC), Melanoma (MEL), Actinic Keratosis (ACK), Seborrheic Keratosis (SEK), and Nevus (NEV). Diagnostic labels are defined at the lesion level and were established by either biopsy or clinical consensus, as provided by the original dataset.

We performed five-fold cross-validation with patient-level partitioning. All lesions and images belonging to the same patient were assigned to the same fold, and no patient was shared between any two folds. In each cross-validation round, four folds were used for training and the remaining fold was held out for validation. Table~\ref{tab:pad_fold_distribution} summarizes the patient-, lesion-, and class-level distributions.

\begin{table}[ht]
\centering
\scriptsize
\setlength{\tabcolsep}{2.2pt}
\renewcommand{\arraystretch}{1.05}
\begin{tabular}{@{}c rr rrrrrr@{}}
\toprule
Fold & Pats & Les & BCC & SCC & MEL & ACK & SEK & NEV \\
\midrule
1 & 275 & 379 & 130 & 25 &  6 & 120 & 45 & 53 \\
2 & 274 & 378 & 121 & 32 &  3 & 133 & 42 & 47 \\
3 & 274 & 378 & 131 & 27 &  9 & 140 & 30 & 41 \\
4 & 275 & 378 & 132 & 31 & 10 & 126 & 40 & 39 \\
5 & 275 & 378 & 138 & 30 &  8 & 125 & 41 & 36 \\
\midrule
All & 1,373 & 1,891 & 652 & 145 & 36 & 644 & 198 & 216 \\
\bottomrule
\end{tabular}
\caption{Patient, lesion, and class distributions across the five
PAD-UFES-20 folds.}
\label{tab:pad_fold_distribution}
\end{table}

Each image was independently processed by the vision encoder. When multiple images were available for the same lesion, their feature vectors were averaged to obtain a single lesion-level visual representation before Stage 3 diagnostic aggregation. The same lesion-level image aggregation was applied to all comparison methods.

PAD-UFES-20 provides 21 metadata attributes covering patient demographics, lifestyle and environmental exposure, medical and family history, and lesion characteristics. Stage 2 used nine lesion-related attributes with potential visual correspondence as semantic anchors: region, horizontal diameter, vertical diameter, itch, growth, pain, change, bleeding, and elevation. Soft consistency loss uses the diagnostic label and seven binary attributes: region, itch, growth, pain, change, bleeding, and elevation. Stage 3 used all 21 metadata attributes for diagnostic aggregation, while the diagnostic label and biopsy-status fields were excluded from the model inputs.

For Stage 2, missing semantic-anchor values were encoded as zero, so they contributed no attribute-level similarity signal and did not introduce artificial correspondence between samples. For Stage 3, structured metadata values were converted into textual descriptions, with missing values retained as explicit unknown statements. Table~\ref{tab:pad_metadata} reports the complete attribute set and the lesion-level missing rate of each attribute.

\begin{table}[ht]
\centering
\scriptsize
\setlength{\tabcolsep}{3.0pt}
\renewcommand{\arraystretch}{1.05}
\begin{tabular}{@{}l r l r@{}}
\toprule
Attribute & Missing (\%) & Attribute & Missing (\%) \\
\midrule
Age                  & 0.00  & Gender                & 37.65 \\
Background father    & 42.09 & Background mother     & 41.78 \\
Smoking              & 37.65 & Alcohol consumption   & 37.65 \\
Pesticide exposure   & 37.65 & Piped-water access    & 37.65 \\
Sewage-system access & 37.65 & Skin-cancer history   & 37.65 \\
Cancer history       & 37.65 & Fitzpatrick skin type & 37.65 \\
Region               & 0.00  & Horizontal diameter   & 37.65 \\
Vertical diameter    & 37.65 & Itch                  & 0.32 \\
Growth               & 20.47 & Pain                  & 0.42 \\
Change               & 20.15 & Bleeding              & 0.32 \\
Elevation            & 0.11  &                       &       \\
\bottomrule
\end{tabular}
\caption{Missing rates of the 21 PAD-UFES-20 metadata attributes, calculated over the 1,891 unique lesions after removing duplicate images of the same lesion.}
\label{tab:pad_metadata}
\end{table}

\subsection{AQUA}
\label{sec:sup_aqua}

AQUA is an institutional slit-lamp photography dataset collected under institutional review board--approved protocols. Site-identifying details are omitted owing to institutional and patient-privacy restrictions. It contains 19,567 images from 1,827 subjects and focuses on binary classification between fungal and bacterial keratitis. For site-stratified analyses, acquisition locations were grouped into India and the United States (US), comprising 1,251 and 576 subjects, respectively.

Ground-truth diagnoses were established through expert adjudication using microbiological culture and PCR-based testing as the primary reference standards. Culture-negative cases were retained when a diagnosis could be established through expert chart review and consultation with the treating clinician. Cases that could not be reliably assigned to either diagnostic category after expert review were excluded.

Up to three SLP illumination modalities were acquired for each subject: diffuse blue-light illumination with topical fluorescein staining (B), sclerotic-scatter illumination (ScS), and diffuse white-light illumination (W). Because AQUA reflects real-world clinical practice, image availability varied across subjects. Some imaging sessions contained multiple images under the same illumination condition, whereas others lacked one or more modalities.

Each available image was independently processed by the vision encoder. When multiple images of the same modality were available for one subject, their feature vectors were averaged to obtain a single subject-level representation for that modality. Each subject was therefore represented by three modality-specific feature slots. When a modality was unavailable, its corresponding feature representation was replaced with a zero vector. The same modality-wise averaging and missing-modality strategy was applied to all multi-modal comparison methods.

We performed five-fold subject-level cross-validation, ensuring that all images belonging to the same subject remained in the same fold. No subject was shared between any two folds. In each cross-validation round, four folds were used for training and the remaining fold was held out for validation. The class distribution differed substantially between the two site groups: the India cohort contained 1,035 fungal and 216 bacterial cases, whereas the US cohort contained 94 fungal and 482 bacterial cases. Table~\ref{tab:aqua_fold_distribution} reports the class-by-site distribution in each fold.

\begin{table}[t]
\centering
\scriptsize
\setlength{\tabcolsep}{2.8pt}
\renewcommand{\arraystretch}{1.05}
\begin{tabular}{@{}c rr rr r@{}}
\toprule
& \multicolumn{2}{c}{India} &
  \multicolumn{2}{c}{US} & \\
\cmidrule(lr){2-3}\cmidrule(lr){4-5}
Fold & Fung. & Bact. & Fung. & Bact. & Total \\
\midrule
1 & 212 & 44 & 10 &  99 & 365 \\
2 & 210 & 44 & 24 &  87 & 365 \\
3 & 193 & 41 & 22 & 110 & 366 \\
4 & 211 & 31 & 18 & 105 & 365 \\
5 & 209 & 56 & 20 &  81 & 366 \\
\midrule
Total & 1,035 & 216 & 94 & 482 & 1,827 \\
\bottomrule
\end{tabular}
\caption{Class-by-site distributions across the five AQUA folds.
Fung. and Bact. denote fungal and bacterial keratitis, respectively.}
\label{tab:aqua_fold_distribution}
\end{table}

AQUA contains 37 metadata attributes spanning patient demographics, environmental and behavioral risk factors, symptoms, ocular history and visual function, ocular examination findings, and free-text corneal examination notes. All metadata attributes and free-text notes were manually reviewed, and all diagnostic expressions, microorganism names, microbiological results, and other target-revealing information were removed to minimize the risk of data leakage.

Stage 2 used seven visually observable ocular examination findings as semantic anchors: granular stromal infiltrate, suppurative stromal infiltrate, feathery stromal infiltrate, purulent stromal infiltrate, smooth ring-shaped stromal infiltrate, hypopyon, and satellite lesions. Soft consistency loss uses the diagnostic label and all the seven binary attributes listed above. Stage 3 used all 37 metadata attributes, including the cleaned corneal examination note, for diagnostic aggregation.

For Stage 2, missing semantic-anchor values were encoded as zero so that they contributed no attribute-level similarity signal. For Stage 3, missing structured metadata were represented using explicit unknown textual descriptions, while an unavailable corneal examination note was represented by an empty string. Because metadata availability differed substantially between the two site groups, Table~\ref{tab:aqua_metadata} reports site-stratified as well as overall missing rates.

\begin{table}[t]
\centering
\scriptsize
\setlength{\tabcolsep}{2.4pt}
\renewcommand{\arraystretch}{1.00}
\begin{tabular}{@{}p{0.50\columnwidth} r r r@{}}
\toprule
Attribute & India & US & Overall \\
\midrule
Corneal exam note                  & 0.40   & 12.15 & 4.11 \\
Contact-lens use                   & 0.00   & 2.95  & 0.93 \\
Age                                & 0.00   & 0.35  & 0.11 \\
Diabetes                           & 0.08   & 0.35  & 0.16 \\
Corneal disease                    & 0.00   & 0.00  & 0.00 \\
Granular infiltrate                & 0.00   & 52.08 & 16.42 \\
Suppurative infiltrate             & 0.00   & 52.08 & 16.42 \\
Feathery infiltrate                & 0.00   & 52.08 & 16.42 \\
Purulent infiltrate                & 0.00   & 52.08 & 16.42 \\
Ring-shaped infiltrate             & 0.00   & 52.08 & 16.42 \\
Hypopyon                           & 0.00   & 53.30 & 16.80 \\
Satellite lesions                  & 0.00   & 52.26 & 16.48 \\
Corrected visual acuity            & 0.00   & 50.69 & 15.98 \\
Uncorrected visual acuity          & 0.00   & 34.90 & 11.00 \\
Trauma                             & 0.00   & 52.60 & 16.58 \\
Pain or discomfort                 & 0.00   & 51.39 & 16.20 \\
Redness                            & 0.00   & 51.39 & 16.20 \\
Photophobia                        & 0.00   & 51.39 & 16.20 \\
Blurry vision                      & 0.00   & 51.39 & 16.20 \\
Outdoor occupation                 & 0.00   & 57.99 & 18.28 \\
Glaucoma                           & 0.00   & 0.00  & 0.00 \\
Vision-impairing eye disease       & 0.00   & 0.00  & 0.00 \\
Sex                                & 0.16   & 0.17  & 0.16 \\
Tobacco history                    & 0.00   & 52.26 & 16.48 \\
Complaint of redness               & 0.00   & 73.96 & 23.32 \\
Degree of thinning                 & 0.00   & 74.31 & 23.43 \\
Conjunctival-injection degree      & 0.00   & 73.96 & 23.32 \\
Central-corneal ulcer              & 0.00   & 74.31 & 23.43 \\
Water exposure                     & 0.96   & 73.96 & 23.97 \\
Recent contact-lens use            & 2.56   & 73.96 & 25.07 \\
Race                               & 0.00   & 2.60  & 0.82 \\
Geography                          & 0.00   & 0.00  & 0.00 \\
Hypopyon height                    & 51.32  & 84.03 & 61.63 \\
Organic-matter exposure            & 28.78  & 93.06 & 49.04 \\
Metallic foreign body              & 28.78  & 93.06 & 49.04 \\
Medical insurance                  & 100.00 & 52.08 & 84.89 \\
Corneal ulcer                      & 100.00 & 26.04 & 76.68 \\
\bottomrule
\end{tabular}
\caption{Site-stratified missing rates (\%) of the 37 AQUA metadata
attributes. India and US include 1,251 and 576 subjects, respectively;
overall rates are calculated across all 1,827 subjects.}
\label{tab:aqua_metadata}
\end{table}

\subsection{Common Training Evaluation Protocol}
\label{sec:sup_protocol}

For both PAD-UFES-20 and AQUA, training images were augmented using random cropping, flipping, rotation, and color jittering. Detailed augmentation parameters are provided in the released code. Validation images were not augmented, and all images underwent the same standard normalization procedure before being passed to the vision encoder.

In each cross-validation round, the checkpoint achieving the highest F1-score on the held-out validation fold was retained, and the corresponding validation metrics were reported. All comparison methods used the same cross-validation folds, image preprocessing, and multi-image aggregation rules. Predictions were obtained by selecting the diagnostic category with the largest output logit; no decision-threshold tuning was performed. We report F1-score, accuracy, and balanced accuracy (BAcc). The reported mean and standard deviation were computed across the five held-out validation folds.

\subsection{Data Integrity and Foundation-Model Pretraining}
\label{sec:sup_pretraining_exposure}

The patient-level partitions prevent direct overlap between the training and held-out data used in the PRIMA experiments. Identical folds were used for all comparison methods. No held-out image or patient record was used for task-specific pretraining and knowledge-corpus construction.

A separate concern is whether the publicly released pretrained parameters adopted by PRIMA were exposed to the evaluation datasets during the original training of their foundation models. The reported pretraining data of DINOv3 and Clinical ModernBERT do not include the datasets evaluated in this study. Although the complete pretraining corpus of Qwen3 is not publicly disclosed, within our framework Qwen3 receives neither raw evaluation images nor original patient records, metadata, or reports. Instead, it processes continuous image and metadata features generated by our encoders and task-specific alignment modules. Thus, even if Qwen3 had encountered related public medical content, it would not directly access the evaluated samples in their original image or text form.

Moreover, Qwen3 operating on features without PRIMA alignment performed substantially worse than the complete framework, as shown in the ablation study. The improvement emerged only after task-specific knowledge pretraining and multi-granular image--metadata alignment, indicating that the diagnostic gain derives from PRIMA-learned representations rather than direct memorization by Qwen3. Therefore, although Qwen3's full original pretraining corpus cannot be independently audited, we find no evidence that foundation-model exposure caused leakage in our experiments.


\section{Implementation and Reproducibility Details}
\label{sec:sup_implementation}

\subsection{Knowledge Corpus Construction}
\label{sec:sup_knowledge}

We constructed a document-grounded knowledge corpus describing the clinical associations between metadata-derived factors and the diagnostic categories of each dataset. The corpus covers 21 factors for PAD-UFES-20 and 37 factors for AQUA. The grounding collections comprised 25 PubMed-indexed publications on dermatology and skin oncology for PAD-UFES-20 and 25 PubMed-indexed publications on infectious ophthalmology and microbial keratitis for AQUA. These collections included clinical guidelines, reviews, cohort studies, epidemiological studies, and diagnostic studies.

The publications were uploaded to the native document-grounding interfaces of ChatGPT and Gemini, which internally retrieved relevant content during generation. PAD-UFES-20 descriptions were independently generated using GPT-5.1 and Gemini 3 Pro, and AQUA descriptions were generated using GPT-5 and Gemini 2.5 Pro. All models used the default settings of their respective interfaces; no decoding parameters were manually specified.

For each metadata factor, the models generated an overview followed by diagnosis-specific clinical descriptions. PAD-UFES-20 responses contained six diagnosis-specific paragraphs covering Basal Cell Carcinoma (BCC), Squamous Cell Carcinoma (SCC), Actinic Keratosis (ACK), Seborrheic Keratosis (SEK), Melanoma (MEL), and Nevus (NEV). AQUA responses contained two subtype-specific paragraphs covering bacterial and fungal keratitis. The prompts required the generated content to remain grounded in the uploaded medical literature, avoid unsupported associations, and explicitly acknowledge insufficient evidence when applicable.

For PAD-UFES-20, the generated corpus contained 42 model responses, which were divided into 294 structured text segments totaling approximately 46,298 words. For AQUA, it contained 74 model responses and 222 structured segments totaling approximately 31,566 words. Each segment was treated as an independent knowledge statement for Stage-1 masked-language-modeling pretraining, and the samples from each dataset were randomly divided into 90\% training and 10\% validation subsets.

The outputs from both models were retained independently and jointly used for Stage-1 text-encoder training. We used the same general prompt structure for both datasets. The placeholders in Listing~\ref{lst:knowledge_prompt} were replaced using the dataset-specific settings summarized in Table~\ref{tab:rag_settings}. For PAD-UFES-20, one diagnosis-specific paragraph was requested for each of the six skin-lesion categories; for AQUA, one paragraph was requested for each keratitis subtype.

\begin{listing}[htbp]
\begin{lstlisting}
RISK_FACTOR = {RISK_FACTOR}. You are a medical researcher specializing in {CLINICAL_DOMAIN}. You are given a specific clinical or metadata-derived factor: {RISK_FACTOR}. Your task is to generate a structured description of the relationship between this factor and the following diagnostic categories: {DIAGNOSTIC_CATEGORIES}. Base the response only on the medical literature provided through the document-grounded interface. Follow this format strictly:

1. Overview Sentence
Provide one or two sentences summarizing the general clinical relationship between {RISK_FACTOR} and the diagnostic categories.

2. Diagnosis-Specific Descriptions
For each diagnostic category, provide one detailed paragraph describing how {RISK_FACTOR} relates to disease risk, occurrence, clinical presentation, severity, progression, or prognosis, as applicable.

Important requirements:

- Use professional academic medical language.
- Ground all statements in the supplied clinical literature.
- Include epidemiological findings, biological mechanisms, clinical observations, or representative statistics only when supported by the supplied documents.
- Do not introduce unsupported facts or associations.
- If the supplied literature does not establish an association, state this explicitly.
- Produce coherent text suitable for training a BERT-based clinical text encoder.
\end{lstlisting}
\caption{Shared prompt template used for document-grounded knowledge
generation.}
\label{lst:knowledge_prompt}
\end{listing}

\begin{table}[ht]
\centering
\small
\setlength{\tabcolsep}{4pt}
\begin{tabularx}{\columnwidth}{@{}p{0.27\columnwidth}X@{}}
\toprule
\textbf{Setting} & \textbf{Configuration} \\
\midrule

PAD-UFES-20 &
21 metadata-derived factors; 25 PubMed-indexed publications;
dermatology and skin oncology; GPT-5.1 and Gemini 3 Pro;
one overview and six descriptions for BCC, SCC, ACK, SEK,
MEL, and NEV. \\

AQUA &
37 metadata-derived factors; 25 PubMed-indexed publications;
infectious ophthalmology and microbial keratitis; GPT-5 and
Gemini 2.5 Pro; one overview and two descriptions for
bacterial and fungal keratitis. \\

Document grounding &
Native file-grounding interfaces of ChatGPT and Gemini. \\

Source types &
Clinical guidelines, reviews, cohort studies, epidemiological
studies, and diagnostic studies. \\

Generation settings &
Default model-interface settings; no decoding parameters were
manually specified. \\

Corpus use &
Both model outputs were retained independently and jointly
used for Stage-1 text-encoder training. \\

Source availability &
The grounding publications are not redistributed because they
remain subject to publisher copyright restrictions. \\

\bottomrule
\end{tabularx}
\caption{Dataset-specific document-grounded knowledge-generation
settings.}
\label{tab:rag_settings}
\end{table}

\subsection{Optimization and Model Selection}
\label{sec:sup_optimization}

All experiments were conducted on a workstation equipped with two NVIDIA RTX~4090 GPUs, each with 24\,GB of memory. AdamW was used throughout the framework. Mixed-precision training was used and gradient norms were clipped to 1.0.

Stage~1 was trained on the independently constructed knowledge bank and did not use any patient images or patient metadata. The knowledge samples were randomly divided into 90\% training and 10\% validation subsets. We used a per-device batch size of 16, eight gradient-accumulation steps, a learning rate of $5\times10^{-5}$, a weight decay of $10^{-3}$, and a maximum of 2500 epochs. Training used FP16 precision, and the learning-rate scheduler was left at the default setting of the Hugging Face Trainer. The checkpoint with the highest validation masked-language-modeling accuracy was retained. Because Stage 1 used only literature-derived, dataset-schema-level knowledge statements and did not consume any patient records, images, labels, or fold-specific statistics, the same Stage~1 encoder was reused across cross-validation rounds. 

Stage~2 comprised alignment pre-training and supervised fine-tuning on the same patient-level folds. Both phases used a batch size of 32, a weight decay of $10^{-3}$, cosine learning-rate decay, and at most 150 epochs. Alignment pre-training used a learning rate of $3\times10^{-5}$ with 15 warm-up epochs and selected the checkpoint with the lowest validation alignment loss, whereas supervised fine-tuning used a learning rate of $1\times10^{-5}$ with 20 warm-up epochs and selected the checkpoint with the highest validation macro F1-score for feature extraction.

Stage~3 used the same patient-level folds as Stage~2. It was trained with a batch size of 8, a learning rate of $1\times10^{-5}$, a weight decay of $3\times10^{-2}$, and a maximum of 80 epochs. Cosine learning-rate decay was used with 10 warm-up epochs. The checkpoint with the highest validation macro F1-score was used for evaluation.

\subsection{Model Components and Parameter-Efficient Adaptation}
\label{sec:sup_model_configuration}

The image encoder was initialized from the DINOv3 ViT-B/16 checkpoint \url{https://huggingface.co/facebook/dinov3-vitb16-pretrain-lvd1689m}. The clinical text encoder was initialized from \url{https://huggingface.co/Simonlee711/Clinical_ModernBERT}.
Stage~3 used \url{https://huggingface.co/Qwen/Qwen3-1.7B}.

LoRA was used to adapt the three pretrained backbones while preserving their original parameters. In Stage~1, LoRA was applied to the query and value projections of the Clinical ModernBERT self-attention layers, with rank 8, scaling factor 8, and dropout 0.2. In Stage~2, LoRA was applied to the query, key, and value projections of DINOv3, with rank 32, scaling factor 32, and dropout 0.2. The knowledge-refined Clinical ModernBERT encoder was frozen during image--metadata alignment. The newly introduced image and text projection heads, fusion module, normalization layers, and diagnostic classifier remained trainable. The temperature $\tau$ and $\tau_{\text{label}}$ are 0.07 and 0.05, respectively.

In Stage~3, LoRA was applied to the Qwen3 attention projections (\texttt{q\_proj}, \texttt{k\_proj}, \texttt{v\_proj}, and \texttt{o\_proj}) and feed-forward projections (\texttt{gate\_proj}, \texttt{up\_proj}, and \texttt{down\_proj}). The LoRA rank, scaling factor, and dropout were 16, 16, and 0.2, respectively. The original Qwen3 token-embedding matrix was frozen, whereas the image and metadata feature projectors were optimized jointly with the LoRA parameters.

\subsection{Feature and Token Construction}
\label{sec:sup_token_construction}

For a $224\times224$ input image, DINOv3 produced one global CLS token and 196 patch tokens arranged on a $14\times14$ spatial grid. Register tokens, when present in the pretrained architecture, were excluded from the patch sequence. During Stage~2, both visual and textual representations were projected into a shared 768-dimensional space. For Stage~3, the aligned features were mapped into the Qwen3 hidden space. The 196 visual patch tokens were reshaped into a $14\times14$ grid and passed through a two-dimensional convolution. This operation produced a $7\times7$ grid, corresponding to 49 local visual tokens per available image modality. The metadata sequence was compressed using a one-dimensional convolution. A 512-token metadata sequence therefore produced 128 projected metadata tokens. All projected tokens had the same dimension as the Qwen3 hidden representations.

The resulting embeddings were concatenated in the following order:
\[
\begin{aligned}
\mathbf{X}=[
&\texttt{<|img\_start|>},\mathbf{v}_{\mathrm{cls}},
\mathbf{V},\texttt{<|img\_end|>},\\
&\texttt{<|txt\_start|>},\mathbf{T},
\texttt{<|txt\_end|>},\mathbf{P} ].
\end{aligned}
\]
where $\mathbf{V}=\{\mathbf{v}_k\}_{k=1}^{K}$, $\mathbf{T}=\{\mathbf{t}_l\}_{l=1}^{L}$, and $\mathbf{P}=\{\mathbf{p}_p\}_{p=1}^{P}$ denote the local visual tokens, metadata tokens, and fixed prompt embeddings, respectively. Four boundary tokens were added to the tokenizer: \texttt{<|img\_start|>}, \texttt{<|img\_end|>}, \texttt{<|txt\_start|>}, and \texttt{<|txt\_end|>}. Their embeddings were initialized from the existing Qwen3 embeddings of the words "image", "text", and "end", and were subsequently frozen together with the original embedding matrix.

\subsection{Vocabulary-Restricted Diagnostic Prediction}
\label{sec:sup_restricted_prediction}

Stage~3 treated Qwen3 as a vocabulary-restricted classifier rather than an open-ended text generator. The prompt listed the complete set of diagnostic categories and ended with an answer prefix. Each diagnostic category was tokenized with a preceding whitespace, and the first resulting vocabulary token was used as its class token.

Let $\mathcal{C}$ denote the predefined diagnostic vocabulary and let $z_{c}$ be the final-position Qwen3 logit associated with the class token of category $c\in\mathcal{C}$. The restricted class probability was computed as
\[
p(c\mid\mathbf{x})
=
\frac{\exp(z_c)}
{\sum_{c'\in\mathcal{C}}\exp(z_{c'})}.
\]
Cross-entropy loss was applied only to these restricted class logits. The final diagnosis was
\[
\hat{c}
=
\arg\max_{c\in\mathcal{C}} z_c.
\]
Consequently, Qwen3 could select only one of the predefined diagnostic categories and could not produce an unrestricted free-text answer.

\subsection{Image and Text Preprocessing}
\label{sec:sup_img_text_processing}

All images were resized to $224\times224$ and normalized using the ImageNet channel statistics $\mu=(0.485,0.456,0.406), \sigma=(0.229,0.224,0.225).$ During training, color jittering was first applied and was followed by horizontal and vertical flipping, each with probability 0.5; random rotation within $[-30^\circ,30^\circ]$; and random resized cropping with a scale range of $[0.8,1.0]$ and an aspect-ratio range of $[3/4,4/3]$. Hue jittering was not used.  Validation and evaluation images were resized and normalized without random augmentation.

Stage~1 used the Clinical ModernBERT tokenizer to encode paired knowledge statements. Inputs were truncated to at most 512 tokens and dynamically padded within each batch. Masked-language-modeling corruption was applied with a masking probability of 0.15. For patient metadata, the textual descriptions of the selected attributes were concatenated. The tokenizer-specific special tokens were added automatically. Fixed-length metadata features used by Stage~3 were padded or truncated to 512 tokens before extraction from the final hidden layer of the knowledge-refined Clinical ModernBERT encoder.

\subsection{Code and Reproducibility Package}
\label{sec:sup_code}

Our code is available at \url{https://github.com/yqwang01/PRIMA}. The repository contains the knowledge-pretraining pipeline, data-loading interfaces, patient-level fold definitions, image--metadata alignment objectives, feature-extraction scripts, vocabulary-restricted Qwen3 classifier, evaluation scripts, and inference procedures required to reproduce the reported experiments. The principal software environment used Python~3.11.11, CUDA~12.1, PyTorch~2.5.1, Torchvision~0.20.1, Transformers~5.1.0, PEFT~0.18.1, Accelerate~1.9.0, and scikit-learn~1.6.1.

PAD-UFES-20 is publicly available under the CC BY 4.0 license from
\url{https://data.mendeley.com/datasets/zr7vgbcyr2/1} (DOI: \texttt{10.17632/zr7vgbcyr2.1}).
The private AQUA dataset is not redistributed because of institutional
restrictions and patient-privacy requirements.

\section{Additional Experiments}
\label{sec:sup_additional_experiments}

\subsection{Hyperparameter Sensitivity}
\label{sec:sup_hyperparameter_sensitivity}

We further examine the sensitivity of PRIMA to its principal hyperparameters on PAD-UFES-20. Unless otherwise specified, only one hyperparameter was varied at a time, while the remaining settings were fixed to the values used in the main experiments. 

\subsubsection{Sensitivity to Loss Weights}
\label{sec:ablation_loss_weights}

We also analyze the sensitivity of PRIMA to the loss weights $\beta_1$--$\beta_4$ and the disease upweighting factor used in $\mathcal{L}_{soft}$. These hyperparameters were selected using validation performance within the five-fold cross-validation protocol. As shown in Table~\ref{tab:ablation_beta}, PRIMA is not overly sensitive to the exact loss-weight configuration. Uniform weights achieve 71.63\% average F1-score and 77.05\% accuracy, which are close to the final setting. The alternative weighting strategy also yields comparable balanced accuracy. The final configuration $(0.2, 0.3, 0.2, 0.3)$ obtains the best average F1-score and accuracy, suggesting a mild benefit from assigning slightly larger weights to the semantic alignment terms.

\begin{table}[ht]
\centering
\small
{
\setlength{\tabcolsep}{4pt}
\begin{tabular}{lccc}
\toprule
$\boldsymbol{(\beta_1,\beta_2,\beta_3,\beta_4)}$
& \textbf{Avg. F1}
& \textbf{Acc.}
& \textbf{BAcc.} \\
\midrule
$(0.25,\,0.25,\,0.25,\,0.25)$
& 71.63
& 77.05
& 71.50 \\

$(0.30,\,0.20,\,0.30,\,0.20)$
& 71.38
& 75.46
& \textbf{72.03} \\

$(0.20,\,0.30,\,0.20,\,0.30)$
& \textbf{72.04}
& \textbf{77.26}
& 71.95 \\
\bottomrule
\end{tabular}
}
\caption{Sensitivity analysis of the loss weights
$\beta_1$--$\beta_4$ on PAD-UFES-20 (\%).}
\label{tab:ablation_beta}
\end{table}

\subsubsection{Sensitivity to Soft-Target Upweighting}
\label{sec:ablation_upweighting}

\begin{table}[ht]
\centering
\small
{
\setlength{\tabcolsep}{5pt}
\begin{tabular}{lccc}
\toprule
\textbf{Upweighting Factor}
& \textbf{Avg. F1}
& \textbf{Acc.}
& \textbf{BAcc.} \\
\midrule
$1$
& 70.06
& 75.99
& 70.68 \\

$2$
& 71.68
& 76.52
& \textbf{73.84} \\

$3$
& \textbf{72.04}
& \textbf{77.26}
& 71.95 \\
\bottomrule
\end{tabular}
}
\caption{Sensitivity analysis of disease upweighting in
$\mathcal{L}_{\mathrm{soft}}$ on PAD-UFES-20 (\%).}
\label{tab:ablation_upweight}
\end{table}

We further evaluate the disease upweighting factor in $\mathcal{L}_{soft}$, which balances the disease-level soft target against multiple metadata attributes. As shown in Table~\ref{tab:ablation_upweight}, using factors of 1, 2, and 3 all yields competitive performance. Increasing the factor from 1 to 3 improves the average F1-score from 70.06\% to 72.04\% and accuracy from 75.99\% to 77.26\%. Although the factor of 2 achieves the highest balanced accuracy, the factor of 3 provides the best average F1-score and accuracy, and is therefore used as the default setting. These results indicate that the proposed metadata-aware soft supervision is robust to moderate changes in its weighting scheme.

\subsection{Efficiency Analysis}
\label{sec:sup_efficiency}

We report both training efficiency and inference cost. Stage~1 employs Clinical ModernBERT with rank-8 LoRA, resulting in 137.17M total parameters, of which only 0.54M (0.39\%) are trainable. Processing one 512-token image-associated metadata sequence requires approximately 150.9 GFLOPs, 0.32 GiB of peak GPU memory, and 11.28 ms. Stage~2 combines DINOv3 ViT-B/16, the frozen knowledge-enhanced text encoder, and the alignment heads, yielding 236.43M total and 13.60M trainable parameters (5.75\%). Its single-image feature-extraction path at \(224\times224\) resolution requires approximately 36.8 GFLOPs, 0.47 GiB, and 7.53 ms per image.

Stage~3 uses Qwen3-1.7B together with multi-modal projectors and rank-16 LoRA, containing 1.756B total and 35.80M trainable parameters (2.04\%). In the single-image setting, the 197 image tokens are compressed into 50 visual tokens. Together with the compressed 128-token text representation, four modality-boundary tokens, and the 56-token classification prompt, the effective LLM input contains 238 tokens. This setting requires approximately 834.2 GFLOPs, 3.41 GiB of peak GPU memory, and 38.45 ms per case. In the three-image setting, the three 197-token image representations are compressed into 150 visual tokens, increasing the effective LLM input length to 338 tokens. This setting requires approximately 1.193 TFLOPs, 3.47 GiB, and 54.97 ms per case.

When the complete pipeline is executed sequentially, the single-image setting requires approximately 1.02 TFLOPs and 57.25 ms per case. The three-image setting, including three Stage~2 image-encoder forward passes, requires approximately 1.45 TFLOPs and 88.84 ms per case. Peak memory is dominated by Stage~3 and reaches 3.41 GiB and 3.47 GiB for the single- and three-image settings, respectively. Overall, parameter-efficient adaptation restricts optimization to a small fraction of the model parameters while maintaining practical inference costs under both single- and multi-image settings. The Stage~1 training procedure itself is performed only once and is not part of inference. The reported Stage~1 inference cost corresponds to one forward pass through the resulting knowledge-refined metadata encoder.

\begin{table}[t]
\centering
\scriptsize
\setlength{\tabcolsep}{1.5pt}
\renewcommand{\arraystretch}{1.08}
\begin{tabular}{@{}lrrrrr@{}}
\toprule
\textbf{Component} &
\textbf{Total} &
\shortstack{\textbf{Trainable}\\\textbf{Params.}} &
\textbf{FLOPs} &
\shortstack{\textbf{Peak Mem.}\\\textbf{(GiB)}} &
\shortstack{\textbf{Latency}\\\textbf{(ms)}} \\
\midrule
Stage~1
& 137.17M & 0.54M (0.39\%) & 150.9G & 0.32 & 11.28$^\dagger$ \\

Stage~2
& 236.43M & 13.60M (5.75\%) & 36.8G & 0.47 & 7.53 \\

Stage~3 (1 img.)
& 1.756B & 35.80M (2.04\%) & 834.2G & 3.41 & 38.45 \\

Stage~3 (3 imgs.)
& 1.756B & 35.80M (2.04\%) & 1.193T & 3.47 & 54.97 \\
\midrule
E2E (1 img.)
& -- & -- & 1.02T & 3.41 & 57.25 \\

E2E (3 imgs.)
& -- & -- & 1.45T & 3.47 & 88.84 \\
\bottomrule
\end{tabular}
\caption{Training efficiency and inference cost of PRIMA under single- and three-image settings. Stage~2 results are reported per image. $^\dagger$This value denotes one metadata-encoder forward pass and does not represent Stage~1 training during inference.}
\label{tab:efficiency}
\end{table}

\subsection{Statistical Robustness}
\label{sec:sup_statistics}

Statistical comparisons were performed using paired fold-level results because all methods were evaluated on the same five patient-level cross-validation partitions. We used one-sided exact Wilcoxon signed-rank tests to evaluate the prespecified alternative that PRIMA achieved higher performance than each comparator. Zero paired differences, if present, were omitted using the \texttt{wilcox} convention. We additionally report the mean paired difference and rank-biserial correlation ($r_{\mathrm{rb}}$) to describe the magnitude and directional consistency of the fold-level differences.

With five nonzero paired differences, $p=0.03125$ is the smallest attainable exact one-sided value. Therefore, $p=0.03125$ primarily indicates that the observed differences favored PRIMA in all five folds; it should not, by itself, be interpreted as evidence of a large effect or as strong confirmatory inference.

On PAD-UFES-20, PRIMA showed nominally significant fold-level improvements over DINOv3, metadata fusion, MedKLIP, KnoBo, and MLRG in both average F1-score and accuracy ($p=0.03125$). Against MedBLIP, the accuracy comparison yielded $p=0.03125$, whereas the average F1-score comparison yielded $p=0.09375$.

On AQUA, PRIMA showed nominally significant fold-level improvements over DINOv3, metadata fusion, DeepIK, KnoBo, and MLRG in average F1-score and accuracy (all $p=0.03125$). Against MedBLIP, the corresponding $p$-values were $0.06250$ and $0.06250$, respectively. 

Because the cross-validation training sets overlap, the five folds are not fully independent experimental replications. The reported $p$-values are unadjusted for multiple comparisons and are therefore interpreted as exploratory evidence of consistency across the fixed data partitions rather than definitive evidence of general superiority.

\begin{table*}[t]
\centering
\scriptsize
\setlength{\tabcolsep}{3.2pt}
\renewcommand{\arraystretch}{1.05}

\begin{tabular}{llccc ccc ccc}
\toprule
\textbf{Dataset} &
\textbf{Comparator} &
\multicolumn{3}{c}{\textbf{Average F1-score}} &
\multicolumn{3}{c}{\textbf{Accuracy}} &
\multicolumn{3}{c}{\textbf{Balanced accuracy}} \\
\cmidrule(lr){3-5}
\cmidrule(lr){6-8}
\cmidrule(lr){9-11}
&
&
\textbf{$\Delta$} &
\textbf{$r_{\mathrm{rb}}$} &
\textbf{$p$} &
\textbf{$\Delta$} &
\textbf{$r_{\mathrm{rb}}$} &
\textbf{$p$} &
\textbf{$\Delta$} &
\textbf{$r_{\mathrm{rb}}$} &
\textbf{$p$} \\
\midrule

PAD-UFES-20
& DINOv3
& $+7.04$ & $1.00$ & $0.03125$
& $+6.77$ & $1.00$ & $0.03125$
& $+4.88$ & $1.00$ & $0.03125$ \\

PAD-UFES-20
& Metadata fusion
& $+3.94$ & $1.00$ & $0.03125$
& $+3.12$ & $1.00$ & $0.03125$
& $+2.74$ & $0.87$ & $0.06250$ \\

PAD-UFES-20
& MedKLIP
& $+24.61$ & $1.00$ & $0.03125$
& $+22.37$ & $1.00$ & $0.03125$
& $+20.52$ & $1.00$ & $0.03125$ \\

PAD-UFES-20
& KnoBo
& $+25.43$ & $1.00$ & $0.03125$
& $+20.57$ & $1.00$ & $0.03125$
& $+23.88$ & $1.00$ & $0.03125$ \\

PAD-UFES-20
& MedBLIP
& $+2.49$ & $0.73$ & $0.09375$
& $+4.13$ & $1.00$ & $0.03125$
& $+0.05$ & $0.33$ & $0.31250$ \\

PAD-UFES-20
& MLRG
& $+11.76$ & $1.00$ & $0.03125$
& $+12.86$ & $1.00$ & $0.03125$
& $+11.87$ & $0.87$ & $0.06250$ \\

\midrule

AQUA
& DINOv3
& $+8.24$ & $1.00$ & $0.03125$
& $+7.67$ & $1.00$ & $0.03125$
& $+7.15$ & $1.00$ & $0.03125$ \\

AQUA
& Metadata fusion
& $+2.49$ & $1.00$ & $0.03125$
& $+2.52$ & $1.00$ & $0.03125$
& $+2.24$ & $1.00$ & $0.03125$ \\

AQUA
& DeepIK
& $+6.53$ & $1.00$ & $0.03125$
& $+7.12$ & $1.00$ & $0.03125$
& $+4.29$ & $1.00$ & $0.03125$ \\

AQUA
& KnoBo
& $+7.24$ & $1.00$ & $0.03125$
& $+6.62$ & $1.00$ & $0.03125$
& $+7.64$ & $1.00$ & $0.03125$ \\

AQUA
& MedBLIP
& $+6.96$ & $0.87$ & $0.06250$
& $+7.22$ & $0.87$ & $0.06250$
& $+5.54$ & $0.73$ & $0.09375$ \\

AQUA
& MLRG
& $+6.41$ & $1.00$ & $0.03125$
& $+6.13$ & $1.00$ & $0.03125$
& $+6.25$ & $1.00$ & $0.03125$ \\

\bottomrule
\end{tabular}

\caption{Exploratory paired fold-level comparisons between PRIMA and the reported baselines. Mean paired differences ($\Delta$) are reported in percentage points, with positive values favoring PRIMA.}
\label{tab:statistical_results}
\end{table*}

\section{Additional Discussion and Limitations}
\label{sec:sup_discussion}

\subsection{Clinical Discussion of SCC Performance}
\label{sec:sup_scc}

The SCC class achieved an average F1-score of 36.84\%. Across the five
held-out folds, SCC accounted for 145 of 1,891 lesions (7.67\%), of which 60 were correctly classified. Among the 85 misclassified SCC lesions, 72 were predicted as BCC, 10 as ACK, 1 as SEK, and 2 as NEV, indicating a concentrated SCC--BCC--ACK confusion pattern rather than uniformly distributed errors.

This pattern is clinically plausible. SCC and BCC may share ulceration, telangiectasia, pigmentation, scaling, and other ambiguous surface features, whereas ACK and SCC belong to the same squamous-neoplasia continuum and frequently exhibit overlapping erythematous, scaly, and hyperkeratotic appearances \cite{ryu2018features,neagu2020overlap, ulrich2016spectrum,reinehr2019actinic}. Hyperkeratosis may further obscure signs of deeper invasion, making these conditions difficult to distinguish from surface photographs alone.

Metadata provided complementary information, improving the SCC F1-score from 32.95\% for image-only DINOv3 to 33.56\% with direct metadata fusion and 36.84\% with PRIMA. This suggests that knowledge-enhanced alignment partially reduced the ambiguity, but the available general clinical metadata could not fully substitute for pathology-specific evidence. Accordingly, PRIMA should not be used as a stand-alone system to exclude or diagnose SCC. Further improvement may require additional SCC cases, pathology-linked supervision, dermoscopic or histological information, and cost-sensitive objectives that explicitly prioritize malignant-class errors.

\subsection{Clinical Discussion of Site-Imbalanced Performance on AQUA}
\label{sec:sup_aqua_site_imbalance}

Differentiating bacterial from fungal keratitis using slit-lamp appearance alone is clinically challenging. Although fungal keratitis is classically associated with features such as feathery infiltrate margins, satellite lesions, and a dry or raised surface, these findings are neither consistently present nor specific to fungal infection. In a photographic survey, corneal specialists correctly distinguished bacterial from fungal keratitis in fewer than 70\% of cases, and a subsequent international evaluation similarly demonstrated limited expert performance based on slit-lamp photographs alone~\cite{dalmon2012clinical,redd2022expert}. These findings indicate substantial visual overlap between the two etiologies and motivate the use of microbiological evidence rather than image appearance alone as the diagnostic reference. Corneal scraping, microscopy, and culture remain central components of etiological evaluation, while polymerase chain reaction and other molecular tests can provide complementary evidence when conventional microbiology is inconclusive~\cite{cabrera2022infectious,aao2024bacterial}. AQUA therefore uses a composite reference standard based primarily on culture and PCR testing, supplemented by expert adjudication for culture-negative cases, rather than a label inferred solely from slit-lamp appearance.

Table~\ref{tab:aqua_full} reports both overall and site-stratified performance. For conciseness, the single-modality DINOv3 baselines are reported separately for diffuse blue-light imaging with fluorescein staining (B), sclerotic-scatter imaging (ScS), and diffuse white-light imaging (W). For the three-modality baselines and PRIMA, all available images within each modality were encoded individually and averaged to form one subject-level representation per modality; missing modalities were represented by zero vectors. The three modality-level representations were concatenated for the image-only MLP, while the metadata representation was additionally included for the image--metadata MLP. PRIMA received the same three modality-level representations and metadata representation, but used the downstream vocabulary-restricted LLM classifier for feature aggregation. Thus, the image--metadata MLP and PRIMA constitute an input-matched system-level comparison, whereas the single-modality results serve as descriptive image-only references. All baseline methods report final results using the same three imaging modalities together with metadata as input.

The site-stratified results in Table~\ref{tab:aqua_full} reveal an additional challenge arising from the AQUA cohort composition. Across nearly all methods, fungal cases were classified substantially more accurately in India, whereas bacterial cases were classified more accurately in the US. For example, the three-modality image-only MLP achieved F1-scores of 87.09\% for India-fungal and 90.04\% for US-bacterial, but only 51.89\% for India-bacterial and 51.78\% for US-fungal. Adding metadata increased the corresponding majority site--class F1-scores to 89.44\% and 90.86\%, while the minority site--class F1-scores remained much lower at 54.78\% and 52.63\%. This recurrent pattern across image-only, image--metadata, and previously proposed methods indicates a strong association between site and diagnostic label. Consequently, models may partly identify the clinical site through illumination characteristics, acquisition devices, imaging workflow, or metadata and missingness patterns, and then exploit the site-specific label prevalence as a shortcut.

PRIMA achieved the highest overall macro F1-score, accuracy, and balanced accuracy, reaching 85.22\%, 86.04\%, and 85.40\%, respectively. Relative to the input-matched three-modality image--metadata MLP baseline, these values represent improvements of 2.49, 2.52, and 2.25 percentage points. PRIMA also improved site-level macro F1 from 72.11\% to 73.25\% in India and from 71.75\% to 74.39\% in the US. Its US-fungal F1 increased from 52.63\% to 55.94\%, and its US-bacterial F1 increased from 90.86\% to 92.83\%. In India, PRIMA increased fungal F1 from 89.44\% to 91.13\% and bacterial F1 from 54.78\% to 55.36\%. The latter result remained slightly below the 55.92\% obtained by MedBLIP, showing that PRIMA did not dominate every individual site--class subgroup.

These findings indicate that PRIMA improves the input-matched baseline within both sites and reduces some of the degradation observed in underrepresented site--class combinations. Nevertheless, it does not eliminate the site imbalance: the PRIMA F1-scores for India-bacterial and US-fungal cases remained only 55.36\% and 55.94\%, compared with 91.13\% for India-fungal and 92.83\% for US-bacterial cases. Moreover, within-site evaluation does not remove all site-associated information, because images, acquisition conditions, clinical workflow, metadata values, and metadata missingness may continue to encode the originating site. The AQUA results should therefore be interpreted as a system-level comparison within this site-imbalanced cohort rather than evidence of site-invariant disease recognition.

Accurate and robust differentiation of bacterial and fungal keratitis remains substantially more difficult than the pooled performance alone suggests. External validation, leave-one-site-out evaluation, and datasets containing sufficient bacterial and fungal cases at each site will be needed before clinical deployment can be considered. Future work will investigate supervision and sampling strategies that more directly discourage site-dependent shortcuts and improve robustness for the underrepresented India-bacterial and US-fungal subgroups.

\begin{table*}[t]
\centering
\scriptsize
{
\setlength{\tabcolsep}{3.0pt}
\begin{tabular}{lccccccccc|ccc|c}
\toprule
&
\multicolumn{9}{c|}{\textbf{F1-Score}} &
\multicolumn{3}{c|}{\textbf{Accuracy}} &
\textbf{BAcc.} \\
\textbf{Method}
& Fungal
& Bacterial
& India-F
& India-B
& US-F
& US-B
& India
& US
& Overall
& India
& US
& Overall
& Overall \\
\midrule

DINOv3 LoRA (B) 
& 80.90
& 67.82
& 83.13
& 45.09
& 43.76
& 89.20
& 64.11
& 66.48
& 74.36 (3.56)
& 74.30
& 82.12
& 76.06 (3.80)
& 76.84 (2.56) \\

DINOv3 LoRA (ScS) 
& 82.54
& 71.42
& 85.63
& 49.59
& 47.16
& 87.64
& 67.61
& 67.40
& 76.98 (1.78)
& 77.69
& 80.15
& 78.37 (1.60)
& 78.25 (1.42) \\

DINOv3 LoRA (W) 
& 79.78
& 71.45
& 83.81
& 46.31
& 42.98
& 86.85
& 65.06
& 64.92
& 75.61 (2.62)
& 75.17
& 78.75
& 76.38 (2.86)
& 76.60 (2.09) \\

Multi-modal via MLP (w/o metadata)
& 83.99
& 76.54
& 87.09
& 51.89
& 51.78
& 90.04
& 69.49
& 70.91
& 80.26 (3.84)
& 79.73
& 83.67
& 81.00 (3.83)
& 81.00 (3.09) \\

+ Metadata via MLP
& 86.36
& 79.10
& 89.44
& 54.78
& 52.63
& 90.86
& 72.11
& 71.75
& 82.73 (2.63)
& 82.96
& 84.81
& 83.52 (2.66)
& 83.15 (1.93) \\

\midrule

DeepIK \textcolor{gray}{[NPJ Dig. Med. - 24]}
& 80.84
& 76.54
& 83.40
& 52.76
& 51.42
& 92.08
& 68.08
& 71.75
& 78.69 (3.00)
& 75.46
& 86.51
& 78.93 (2.94)
& 81.11 (2.38) \\

KnoBo \textcolor{gray}{[NIPS-24]}
& 83.54
& 72.42
& 88.02
& 29.59
& 21.30
& 87.31
& 58.80
& 54.31
& 77.98 (6.09)
& 79.70
& 78.61
& 79.42 (5.72)
& 77.76 (5.75) \\

MedBLIP \textcolor{gray}{[ACCV-24]}
& 80.96
& 75.55
& 84.88
& 55.92
& 42.21
& 87.70
& 70.40
& 64.96
& 78.26 (5.13)
& 77.73
& 80.19
& 78.82 (5.37)
& 79.86 (4.34) \\

MLRG \textcolor{gray}{[CVPR-25]}
& 83.35
& 74.28
& 86.76
& 44.13
& 47.20
& 89.20
& 65.45
& 68.20
& 78.81 (2.73)
& 78.78
& 82.23
& 79.91 (2.44)
& 79.15 (2.81) \\

\midrule

\textbf{PRIMA (Ours)}
& \textbf{88.66}
& \textbf{81.78}
& \textbf{91.13}
& \textbf{55.36}
& \textbf{55.94}
& \textbf{92.83}
& \textbf{73.25}
& \textbf{74.39}
& \textbf{85.22} (2.17)
& \textbf{85.23}
& \textbf{87.77}
& \textbf{86.04} (1.82)
& \textbf{85.40} (2.27) \\

\bottomrule
\end{tabular}
}
\caption{Quantitative results on the AQUA dataset (\%). Overall results are reported as mean (standard deviation) over five folds, whereas site-specific results are reported without standard deviations. B, ScS, and W denote diffuse blue-light, sclerotic-scatter, and diffuse white-light imaging, respectively. India-F, India-B, US-F, and US-B denote site-specific fungal and bacterial F1-scores.}
\label{tab:aqua_full}
\end{table*}

\subsection{Limitations}
\label{sec:sup_limitations}

This study has several limitations. PRIMA's efficiency primarily refers to parameter-efficient adaptation and the avoidance of large paired image--report corpora; it does not imply that inference with DINOv3, Clinical ModernBERT, and Qwen3-1.7B is computationally lighter than conventional classifiers. Knowledge-corpus construction also incurs a one-time external computational and financial cost, and the generated corpus has not yet been systematically reviewed by clinical experts. In addition, the reported performance is limited to two metadata-rich datasets. Internal validation may yield optimistic performance estimates, and external testing across broader clinical scenarios remains necessary. Further refinement is also needed to address fairness across sensitive subgroups, including sites and races.

Patient-level partitioning prevented direct overlap between the training and validation data used within our pipeline, and no validation image or patient record was used for task-specific pretraining or metadata construction. However, because PRIMA inherits publicly released Qwen3 parameters and the full Qwen3 pretraining corpus is undisclosed, prior exposure to content related to the evaluation datasets cannot be completely ruled out. Nevertheless, our design partially mitigates this concern: in our framework, Qwen3 receives only continuous representations produced by the task-specific encoders rather than raw evaluation images, metadata, or patient records. Moreover, its substantially weaker performance without PRIMA alignment provides no evidence of direct memorization.

Future work will clinically review and refine the generated knowledge corpus, improve fairness across sites and demographic groups, and extend PRIMA to additional datasets and clinical settings. We will also explore more effective strategies for leveraging readily available public data and curated knowledge bases to improve data efficiency and strengthen performance in specialized and rare clinical scenarios where large-scale annotated datasets are difficult to obtain.


\end{document}